\documentclass[preprint,11pt]{elsarticle}

\usepackage{xcolor}
\usepackage{amsmath,amssymb,amsfonts}
\usepackage{amsthm}
\usepackage{multirow}
\usepackage{subfigure}
\usepackage{booktabs}
\usepackage{hyperref}

\journal{Engineering Applications of Artificial Intelligence}

\begin{document}

\begin{frontmatter}



\title{Explainable Artificial Intelligence Techniques for Irregular Temporal Classification of Multidrug Resistance Acquisition in Intensive Care Unit Patients
}

\author[inst1]{Óscar Escudero-Arnanz\corref{cor1}}
\ead{oscar.escudero@urjc.es}
\cortext[cor1]{Corresponding author}
\author[inst1]{Cristina Soguero-Ruiz}
\ead{cristina.soguero@urjc.es}
\author[inst2]{Joaquín Álvarez-Rodríguez}
\ead{joaquin.alvarez@salud.madrid.org}
\author[inst1]{Antonio G. Marques}
\ead{antonio.garcia.marques@urjc.es}

\affiliation[inst1]{organization={Department of Signal Theory and Communications, Telematics and Computing Systems, Rey Juan Carlos University},
            addressline={Camino del Molino, 5}, 
            city={Fuenlabrada},
            postcode={28942}, 
            state={Madrid},
            country={Spain}}

\affiliation[inst2]{organization={University Hospital of Fuenlabrada},
            addressline={Camino del Molino, 2}, 
            city={Fuenlabrada},
            postcode={28942}, 
            state={Madrid},
            country={Spain}}

\begin{abstract}

Antimicrobial Resistance represents a significant challenge in the Intensive Care Unit (ICU), where patients are at heightened risk of Multidrug-Resistant (MDR) infections—pathogens resistant to multiple antimicrobial agents. This study introduces a novel methodology that integrates Gated Recurrent Units (GRUs) with advanced intrinsic and post-hoc interpretability techniques for detecting the onset of MDR in patients across \textit{time}.  Within interpretability methods, we propose Explainable Artificial Intelligence (XAI) approaches to handle \textit{irregular Multivariate Time Series} (MTS), introducing Irregular Time Shapley Additive Explanations (IT-SHAP), a modification of Shapley Additive Explanations designed for irregular MTS with Recurrent Neural Networks focused on temporal outputs. Our methodology aims to identify specific risk factors associated with MDR in ICU patients. GRU with Hadamard's attention demonstrated high initial specificity and increasing sensitivity over time, correlating with increased nosocomial infection risks during prolonged ICU stays. XAI analysis, enhanced by Hadamard attention and IT-SHAP, identified critical factors such as previous non-resistant cultures, specific antibiotic usage patterns, and hospital environment dynamics. These insights suggest that early detection of at-risk patients can inform interventions such as preventive isolation and customized treatments, significantly improving clinical outcomes. The proposed GRU model for \textit{temporal} classification achieved an average Receiver Operating Characteristic Area Under the Curve of 78.27 $\pm$ 1.26 over time, indicating strong predictive performance. In summary, this study highlights the clinical utility of our methodology, which combines predictive accuracy with interpretability, thereby facilitating more effective healthcare interventions by professionals.
\end{abstract}


\end{frontmatter}

\section{Introduction}

In recent years, Artificial Intelligence (AI) has revolutionized various sectors, including healthcare and finance, through the application of Machine Learning (ML) and Deep Learning (DL) models~\cite{AI_health_2023}. These AI systems have demonstrated remarkable capabilities and high performance in diverse scenarios, supporting decision-making across different domains. Despite their outstanding performance, DL models often lack transparency in their decision-making processes, known as the ``black box'' problem, which can affect the reliability and acceptance of these models~\cite{XAI_categories_2022, XAI_taxonomy_2023}. The field of Explainable Artificial Intelligence (XAI) has emerged to address this challenge, focusing on developing AI models that prioritize interpretability alongside performance. XAI encompasses various techniques aimed at making models more understandable and transparent~\cite{XAI_categories_2022, XAI_taxonomy_2023, XAI_health_2023}. This is particularly important in the clinical domain, where accurate predictions must be complemented by interpretability to provide clinicians with insights into relevant variables and potential risk factors~\cite{XAI_categories_2022}.

In the healthcare domain, data from Electronic Health Record (EHR) are essential for addressing numerous current issues~\cite{challenges_MTS_2022}. EHR stores records about patients’ health status and \textit{evolution}, representing \textit{dynamic} information that evolves over time. Given the \textit{temporal} nature of this data, Recurrent Neural Networks (RNNs) are a primary choice in DL architectures to manage these dynamics due to their inherent ability for sequential processing and internal memory mechanisms~\cite{challenges_MTS_2022, analysis_RNN_2023}.

In this context, our research is concentrated on clinical application, with a specific focus on addressing the increasingly concerning issue of Antimicrobial Resistance (AMR), which has profound implications for global economics and demographics~\cite{murray2022global}. Antimicrobials are crucial in the treatment of infectious diseases. However, AMR occurs when pathogens develop resistance to an antimicrobial to which they were previously susceptible~\cite{whoAntimicrobialResistance_2023}. This resistance, caused by genetic alterations in pathogens, reduces the effectiveness of antibiotics and other antimicrobial agents, posing significant challenges in infection management~\cite{whoAntimicrobialResistance_2023}. The acceleration and spread of AMR are primarily due to human activities, particularly the incorrect and excessive use of antimicrobials~\cite{whoAntimicrobialResistance_2023}. The World Health Organization (WHO) has identified AMR as a major public health problem, projecting an increase in healthcare costs of $1$ trillion and losses in Gross Domestic Product of up to $3.4$ trillion annually by 2030, with an estimated 10 million deaths by 2050~\cite{whoAntimicrobialResistance_2023, AMR_deaths_2023}. The COVID-19 pandemic exacerbated this issue in 2020. Uncertainty and lack of information led to the excessive prescription of antibiotics~\cite{covid19_guidelines_2023}. According to the WHO, only 8\% of the hospitalized COVID-19 patients required antibiotics for bacterial infections. However, 75\% of the patients received antibiotics without a confirmed bacterial infection, thereby exacerbating the dissemination of AMR and, consequently, fostering the development of Multidrug Resistant (MDR) pathogens. MDR is characterized by the pathogen's ability to resist multiple antimicrobial agents. A high propagation rate characterizes these MDR microorganisms, significantly reduced treatment options, and elevated mortality rates~\cite{wartu2019multidrug, world2024bacterial}.

Considering the importance of this clinical issue and the relevance of developing explainable architectures in the clinical domain, this study focuses on defining novel XAI strategies combined with RNNs designed for temporal classification. To evaluate these techniques and address the MDR issue, we utilized data from the EHR of patients admitted to the Intensive Care Unit (ICU) of the University Hospital of Fuenlabrada (UHF) in Madrid, Spain. The data, modeled as irregular Multivariate Time Series (MTS), present significant challenges, including varying sampling frequencies, patient records with different lengths, missing data, and heterogeneous features (being a mix of binary, categorical, and numerical data)~\cite{challenges_MTS_2022}. Using these irregular MTS, our objective is to develop an interpretable RNN that allows for analyzing the temporal development of MDR in ICU patients. It also provides clinicians with knowledge about the most relevant variables and risk factors leading to MDR acquisition. The main contributions of this work are as follows\footnote{All experiments and the developed code are publicly accessible in a GitHub repository. The XAI models and RNN developed in this study, along with the corresponding experimental setups, are comprehensively documented and available at the following link: \url{https://github.com/oscarescuderoarnanz/XAI4MTS}.}:
\begin{itemize}
    \item Developing RNNs for the irregular temporal classification of MDR acquisition in ICU patients at UHF.
    \item Adapting XAI models for irregular MTS and RNNs with temporal output, providing clinicians with crucial interpretative tools to enhance decision-making and identify key risk factors and temporal patterns.
    \item Designing a strategy to characterize timeSHAP~\cite{timeshap2021} in scenarios with RNNs featuring temporal output and irregular MTS, enabling precise feature relevance determination at each time step. This method is named Irregular Time Shapley Additive Explanations (IT-SHAP).
    \item Comparing the interpretability and performance of various XAI methods, highlighting their strengths and weaknesses to guide the selection of appropriate techniques for clinical applications.
\end{itemize}

\section{Related Work}

DL models, despite their high performance, often function as ``black boxes" due to their lack of interpretability. This is particularly problematic in clinical settings, especially regarding MDR in critical areas like the ICU, where identifying potential risk factors explaining its spread is essential~\cite{XAI_taxonomy_2023}.

To address this issue, XAI techniques have been developed, which can be grouped into two main categories. The first category is intrinsic explanations, which aim to develop inherently interpretable model architectures. The second category is post-hoc explanations, which focus on the model's output without considering the internal mechanisms of the network~\cite{XAI_categories_2022,XAI_taxonomy_2023}.
Widely used XAI techniques in the clinical domain include SHAP, attention mechanisms, and Feature Selection (FS) methods. SHAP is a post-hoc method that analyzes the model's predictions with and without the presence of a specific feature, estimating the contribution of that particular variable~\cite{shap2017}. Attention mechanisms highlight relevant input variables through an attention module in the DL model architecture~\cite{soydaner2022attention}. Lastly, FS aims to generate a subset of the most relevant features before, during, or after the model training~\cite{figueroa2021towards}.

In the context of MDR detection using EHR modeled as MTS inputs, a study referenced by~\cite{rw_xai4_2020} tested various FS schemes to identify the relevant features and time instants for detection. 
Additionally, in~\cite{rw_xai3_2023_transAMR} explored attention mechanisms, applying a transformer model with an attention module alongside integrated gradients as an interpretability method.
Conversely, other works address the MDR problem without XAI methods. Several recent studies have used various methodologies to enhance the prediction of MDR based on EHR data~\cite{rw_1_2022, rw_2_2023, rw_3_2022, rw_4_2023}. However, these studies often fail to fully exploit the temporal dynamics of the data, relying mainly on ML and DL mechanisms with static variables. For instance, in~\cite{rw_5_2022}, a multivariate dynamic regression model was developed to examine the relationship between different antibiotics and their impact on the development of bacterial resistance to colistin. Nevertheless, these studies tend to focus solely on predicting the resistance of specific pathogens for single outputs, with interpretability not being the primary objective.

The number of studies performing temporal classification using irregular MTS with consistent methodologies is limited. This gap in the literature hinders a deeper understanding of complex temporal patterns. Although the field of forecasting has made significant advances in extracting temporal predictions from sequences, enabling multi-step forecasts in MTS, the specific challenge of classifying irregular MTS remains underexplored. 
For instance, \cite{liu2023time} were able to forecast future values at various time intervals for a train braking system. Similarly, \cite{guo2023multivariate} successfully predicted a week's worth of COVID-19 dynamics. However, there remains a distinct lack of research dedicated to the temporal classification of such irregular data streams~\cite{guo2023multivariate, liu2023time, 2021spatio}.

Members of our research group have been dedicated to the study of MDR in ICUs using data from EHRs since 2019, referencing this issue as AMR in previous papers. Initial studies~\cite{hernandez2020modelling, hernandez2021antimicrobial} 
analyzed the dynamics of \textit{Pseudomonas aeruginosa} considering different predictive time windows. Furthermore, the data began to be modeled as regular MTS, as described in~\cite{martinez2020aplying, escudero2020temporal}, where an RNN model was used to predict MDR, and various FS and ML methods were employed to extract information and predict MDR. In more recent studies~\cite{martinez2022interpretable, martinez2023LSTM}, interpretable XAI techniques were developed for regular MTS and single output approaches. These studies~\cite{martinez2022interpretable, martinez2023LSTM} implemented various FS methods, attention mechanisms, and SHAP. Additionally, they utilized different features and synchronization methods compared to those employed in the MTS of our research.

Given the crucial role of interpretability in the clinical domain, the main contribution of this work is twofold.  On the application side, we are the first work providing time-varying estimates that predict the development of MDR for ICU patients. On the technical side, we develop an XAI approach able to deal with models whose input is an irregular MTS and whose output is a sequence of labels. More specifically, given the popularity of SHAP, we develop IT-SHAP, a variation of SHAP that provides interpretability for models whose inputs are irregular MTS (including RNNs and GRUs) and can be used in applications that go beyond MDR and clinical data~\cite{challenges_MTS_2022, analysis_RNN_2023}.

The remainder of the article is organized as follows. Section~\ref{sec:Methods} presents the notation and methods used in this work. Subsequently, Section~\ref{sec:Database} describes the dataset used. The experiments and results are provided in Section~\ref{sec:DataAnalytic}. Finally, Section~\ref{sec:Disc} presents the main conclusions and discussions.

\section{Methods}\label{sec:Methods}
\begin{figure}[h!]
    \centering
	\centering
	\includegraphics[width=1\columnwidth]{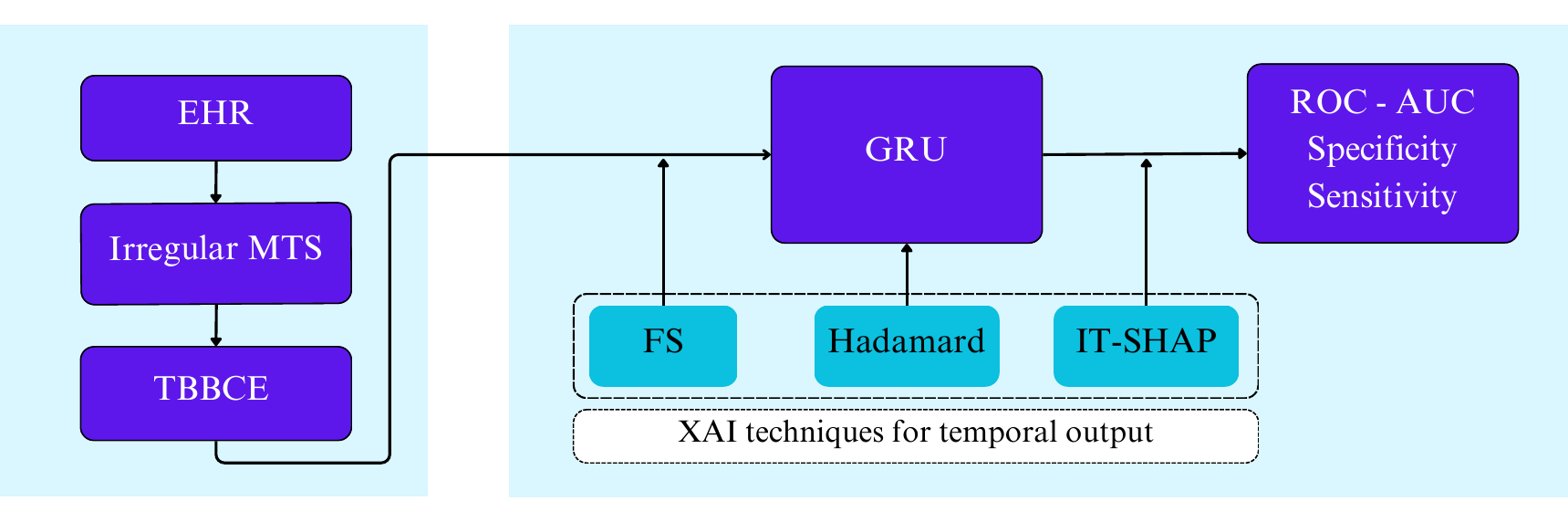}
        \caption{Graphical illustration of the proposed workflow. The process begins with data from Electronic Health Records (EHR), which is modeled into an irregular Multivariate Time Series (MTS). The Temporal Balanced Binary Cross Entropy (TBBCE) is applied to manage class imbalance and temporal dependencies. The pre-processed data is then input into the Gate Recurrent Unit (GRU) model. For interpretability, XAI techniques including Feature Selection (FS), Hadamard, and Irregular Time Shapley Additive Explanations (IT-SHAP) are utilized for temporal outputs. Finally, model performance is evaluated using metrics that include the Receiver Operating Characteristic Area Under the Curve (ROC AUC), Specificity, and Sensitivity.}
	\label{fig:pipeline}
\end{figure}

This section introduces the notation and provides a concise description of the RNNs and explainability methods for irregular temporal classification comprising the architecture proposed in this work, as depicted in Figure~\ref{fig:pipeline}.  

\subsection{Notation}

We represent our patient dataset as $\mathcal{D}=\{(\mathbf{X}_{i}, \mathbf{y}_{i})\}_{i=1}^I$, where $I$ denotes the number of (labeled) patients. The $i$-th patient is represented by the feature matrix $\mathbf{X}_i \in \mathbb{R}^{F \times T}$, where $F$ is the number of features (variables) and $T$ the number of time slots. The $t$-th column of $\mathbf{X}_i$, denoted as $[\mathbf{X}_i]_{(:,t)}$, is a vector that contains the $F$ features of the $i$-th patient for the time slot $t$. Similarly, the $f$-th row of $\mathbf{X}_i$, denoted as $[\mathbf{X}_i]_{(f,:)}$, is the one-dimensional Time Series (TS) representing values of feature $f$ for patient $i$ over time. Patients who have at least one positive MDR culture during their ICU stay are assigned to the MDR class, while all other patients are classified as belonging to the non-MDR class. Given that we are dealing with a temporal binary classification task, we have considered the label ``1" to denote MDR patients and the label ``0" to represent non-MDR patients. Finally, the label for the $i$-th patient is denoted as $\mathbf{y}_i \in \{0,1\}^{T}$, while the hard label estimated (predicted) by the DL model is denoted as $\mathbf{\hat{y}}_i \in \{0,1\}^{T}$.

\subsection{Temporal Classification Model}
\label{Methods_TSA}

RNNs are a class of Neural Networks (NNs) particularly effective for sequential data due to their ability to maintain an internal memory of the previous inputs. Unlike traditional NNs that treat each input independently, RNNs utilize a hidden state to capture dependencies across time steps, making them well-suited for sequential data analysis~\cite{analysis_RNN_2023}. However, standard RNNs suffer from the vanishing gradient problem, where the gradients used in backpropagation diminish exponentially over long sequences, impeding the learning process~\cite{vanishgrad_2021}.

To address the limitations of traditional RNNs, the Gated Recurrent Unit (GRU) was developed as an enhanced variant. GRUs possess a simpler architecture with fewer parameters, leading to faster training times and improved performance compared to conventional RNNs~\cite{GRU_2023}. This is particularly beneficial in medical research, where data can be sparse and computational efficiency is crucial~\cite{challenges_MTS_2022}.

GRU incorporates two primary mechanisms—update and reset gates—to regulate the flow of information. The update gate, denoted as $\mathbf{z}_t$ (\ref{upd_gate}), determines the extent to which past information should be carried forward and combined with new input at each time step. In (\ref{upd_gate}), $\sigma$ represents the sigmoid function, $\mathbf{W}_z$ represents the weights of the update gate, and $\mathbf{b}_z$ is the corresponding bias term~\cite{analysis_RNN_2023}. The reset gate, $\mathbf{r}_t$ (\ref{r_gate}), decides which parts of the past information to discard. $\mathbf{W}_r$ and $\mathbf{b}_r$ are the weights and bias of the reset gate, respectively~\cite{GRU_2023}. The current memory content, or candidate activation, $\widetilde{\mathbf{h}}_t$, is computed using the reset gate (\ref{curr_mem}). In this context, $\tanh$, which represents the hyperbolic tangent function, has been chosen as the activation function, $\mathbf{W}_{\widetilde{\mathbf{h}}}$ and $\mathbf{b}_{\widetilde{\mathbf{h}}}$ are the weights and bias for the candidate activation, and $\odot$ denotes the Hadamard product~\cite{horn1990hadamard}. Finally, the hidden state $\mathbf{h}_t$ (\ref{hid}) is updated by combining the previous hidden state and the candidate activation.

\begin{equation}
    \mathbf{z}_t = \sigma(\mathbf{W}_z \cdot [\mathbf{x}_t, \mathbf{h}_{t-1}] + \mathbf{b}_z)
    \label{upd_gate}
\end{equation}

\begin{equation}
    \mathbf{r}_t = \sigma(\mathbf{W}_r \cdot [\mathbf{x}_t, \mathbf{h}_{t-1}] + \mathbf{b}_r)
    \label{r_gate}
\end{equation}

\begin{equation}
    \widetilde{\mathbf{h}}_t = \tanh(\mathbf{W}_{\widetilde{\mathbf{h}}} \cdot [\mathbf{r}_t \odot \mathbf{h}_{t-1}, \mathbf{x}_t] + \mathbf{b}_{\widetilde{\mathbf{h}}})
    \label{curr_mem}
\end{equation}

\begin{equation}
    \mathbf{h}_t = (1 - \mathbf{z}_t) \odot \widetilde{\mathbf{h}}_t + \mathbf{z}_t \odot \mathbf{h}_{t-1}
    \label{hid}
\end{equation}


The sigmoid activation $\mathbf{z}_t$ modulates the retention of the past hidden state $\mathbf{h}_{t-1}$ based on its value~\cite{GRU_2023}. If $\mathbf{z}_t$ is 1, the hidden state remains unchanged; otherwise, it updates to incorporate the new candidate state based on the current input $\mathbf{x}_t$ and the previous hidden state $\mathbf{h}_{t-1}$. This mechanism allows GRUs to effectively manage the flow of information and maintain relevant historical context, which is crucial for sequential data processing.

As already explained, handling irregularities in MTS data is a significant challenge in medical research. To incorporate this into our architectures, we employ a masking mechanism to ensure that only the available data points are processed by the GRU. Specifically, we use a mask matrix $\mathbf{M}_i \in \{0,1\}^{F \times T}$, where $[\mathbf{M}_{i}]_{(f,t)} = 1$ if the feature $f$ is present at time $t$ for patient $i$, and $0$ otherwise. The masked input $\mathbf{\tilde{X}}_i$ is then computed as $\mathbf{\tilde{X}}_i = \mathbf{X}_i \odot \mathbf{M}_i$, effectively handling the irregularity in the MTS and allowing the GRU to process the data more efficiently.

Our model predicts the probability of MDR at each time step $t$ for each patient $i$. The label for the $i$-th patient is represented as $\mathbf{y}_i \in \{0,1\}^{T}$, while the predicted soft label is $\mathbf{\hat{y}}_i \in \{0,1\}^{T}$. By employing GRUs, we leverage their ability to handle irregular MTS of varying lengths, making them particularly suitable for the temporal analysis of MDR acquisition. This capability, combined with their efficient training and improved performance, underscores the appropriateness of GRUs for our complex medical data analysis tasks.

\subsection{Explainability Methods}
\label{sub:xai}

In the field of XAI, two primary approaches to understanding the decisions made by an AI model exist: intrinsic explainability techniques and post-hoc explainability techniques~\cite{XAI_categories_2022, XAI_health_2023}. Intrinsic techniques integrate interpretability directly into the model design, creating inherently interpretable architectures~\cite{gupta2024comparative}. Conversely, post-hoc techniques explain ``black box" models through various analysis methods after training~\cite{gupta2024comparative}. Intrinsic methods enhance transparency but may reduce accuracy due to design constraints~\cite{ali2023explainable}, while post-hoc methods offer interpretability without significantly affecting performance~\cite{figueroa2021towards}. Thus, a trade-off exists between interpretability and accuracy in XAI techniques~\cite{tradeoff_xai_2020}.

In the healthcare domain, in addition to the two previous approaches, a third one, consisting simply in applying an FS scheme is broadly adopted. Strictly speaking, FS can be classified as a pre-hoc technique, since it is performed before training the model. This contrasts with intrinsic and post-hoc techniques, which are integrated within the model architecture or applied post-training, respectively. The popularity of FS in healthcare applications is partly due to its straightforward interpretation and its ability to incorporate expert domain knowledge.  

In this context, we aim to compare various XAI methods adapted for irregular MTS. Specifically, we focus on the interpretability provided by pre-hoc, intrinsic, and post-hoc techniques and their impact on model performance. Using a GRU model to process irregular MTS for the temporal analysis of MDR acquisition, we adapt these methods for irregular temporal explainability. Additionally, we propose a modification of SHAP for architectures (models) whose inputs are irregular MTS and whose outputs are a sequence (TS) of labels. 

\subsubsection{Pre-Hoc}

FS techniques are essential for refining input data by extracting the most relevant features, improving model interpretability and performance by addressing issues such as overfitting~\cite{XAI_categories_2022}. Among the various FS methods, Conditional Mutual Information (CMI) is one of the most widely used due to its robust theoretical foundation and effectiveness~\cite{gu2022conditional}.

CMI is based on information theory and measures the dependence between two random variables given a third variable, offering a powerful method for FS~\cite{zan2022conditional}. It builds upon the concept of Shannon entropy, which quantifies the uncertainty or probability of occurrence of a random variable $\mathbf{X}$~\cite{MI_shannon_2022, shannon2_2020}. Specifically, the Shannon entropy for a discrete random variable is defined as 
\[
H(\mathbf{X}) = - \sum_{x \in \mathcal{X}} p(x) \log_2 \left( p(x) \right),
\]
where $x$ represents the values of the random variable $\mathbf{X}$, and $p(x)$ is the probability distribution $P\{\mathbf{X} = x\}$. Next, we consider the joint entropy $H(\mathbf{X}, \mathbf{Y})$ of two variables $\mathbf{X}$ and $\mathbf{Y}$, which captures the combined uncertainty of both variables. It is defined as
\[
H(\mathbf{X}, \mathbf{Y}) = - \sum_{x \in \mathcal{X}} \sum_{y \in \mathcal{Y}} p(x, y) \log_2(p(x, y)),
\]
where $p(x, y)$ represents the joint probability $P\{\mathbf{X} = x, \mathbf{Y} = y\}$. To express the uncertainty of one variable given the knowledge of another, we use conditional entropy $H(\mathbf{X} | \mathbf{Y})$, defined as
\[
H(\mathbf{X} | \mathbf{Y}) = - \sum_{x \in \mathcal{X}} \sum_{y \in \mathcal{Y}} p(x, y) \log_2 p(x|y),
\]
where $p(x|y)$ is the conditional probability $P\{\mathbf{X} = x | \mathbf{Y} = y\}$.

The relationship between entropy and mutual information ($I(\mathbf{X}, \mathbf{Y})$), which quantifies the amount of information obtained about one variable through the other, is given by 
\[
I(\mathbf{X}, \mathbf{Y}) = H(\mathbf{X}) - H(\mathbf{X}|\mathbf{Y}) = H(\mathbf{Y}) - H(\mathbf{Y}|\mathbf{X}) = I(\mathbf{Y}, \mathbf{X}).
\]

Finally, we define CMI, which measures the dependence between two random variables given a third variable ($\mathbf{Z}$). CMI is expressed as
\[
I(\mathbf{X}; \mathbf{Y} | \mathbf{Z}) = H(\mathbf{X}, \mathbf{Z}) + H(\mathbf{Y}, \mathbf{Z}) - H(\mathbf{X}, \mathbf{Y}, \mathbf{Z}) - H(\mathbf{Z}).
\]

In the context of FS, CMI is particularly useful for identifying features that maximize the information shared with the class labels $\mathbf{y}$. Our approach applies CMI to all features at each time step $[\mathbf{X}_i]_{(:,t)}$ for temporal FS. The final output for all patients will be $\mathbf{S}_{CMI} \in \mathbb{R}^{F \times T}$. If the value of $[\mathbf{S}_{CMI}]_{(f,t)}$ is close to zero, then the variable $[\mathbf{X}_i]_{(f,t)}$ is deemed as not relevant and can be deselected. In contrast, high values of $[\mathbf{S}_{CMI}]_{(f,t)}$ indicate that the variable is relevant and should be considered.

To account for the nature of our data, which includes both discrete and continuous variables, we adapt the CMI calculation. For discrete variables, probability distributions are estimated directly from the data. For continuous variables, we use kernel density estimation or other methods to approximate the probability densities. Additionally, to handle the irregularity of MTS, we incorporate a masking matrix $\mathbf{M}_i \in \{0,1\}^{F \times T}$ to ignore samples where information is missing due to the irregular nature of the MTS. The masked input $\mathbf{\tilde{X}}_i$ is computed as $\mathbf{\tilde{X}}_i = \mathbf{X}_i \odot \mathbf{M}_i$, ensuring robust and accurate CMI computation. This approach allows effective FS across time steps, enhancing model interpretability and performance by focusing on the most informative features while considering the temporal and irregular nature of the data.

\subsubsection{Intrinsic Methods}

Intrinsic methods aim to enhance interpretability by incorporating techniques directly within the model's architecture. One prominent example is attention mechanisms, which allow the model to focus selectively on critical regions of the input~\cite{XAI_categories_2022, soydaner2022attention}. Originally developed to tackle challenges in machine translation and other natural language processing tasks, attention mechanisms have also been successfully applied in the medical field for tasks such as medical computer vision, image segmentation, classification, and studies in physiology and pharmaceuticals~\cite{attention_2022}.

Our work employs a novel approach by implementing a variable-level attention mechanism that directly operates on the model's input variables~\cite{ICU_30Days_2022, Attention_2019}. In this approach, we use a Gated Residual Network (GRN) to compute an attention matrix \(\mathbf{A}_i \in \mathbb{R}^{F \times T}\) for each patient \(i\), based on the input \(\mathbf{X}_i \in \mathbb{R}^{F \times T}\) and the weights learned by the GRN, utilizing a softmax activation function. Equation~(\ref{att}) represents the computation of the attention matrix \(\mathbf{A}_i\), where \(\mathbf{W}_\text{GRN}\) and \(\mathbf{b}_\text{GRN}\) are the weights and biases of the GRN.

\begin{equation}
    \mathbf{A}_i = \text{softmax}(\mathbf{W}_\text{GRN} \mathbf{X}_i + \mathbf{b}_\text{GRN})
    \label{att}
\end{equation}

The attention matrix $\mathbf{A}_i$ is then element-wise multiplied (Hadamard product) with the original input $\mathbf{X}_i$ to produce a weighted input $\mathbf{X'}_i = \mathbf{X}_i \odot \mathbf{A}_i$. To further refine this, the weighted input $\mathbf{X'}_i$ is multiplied by the mask $\mathbf{M}_i$, resulting in $\mathbf{\tilde{X}}_i = \mathbf{X'}_i \odot \mathbf{M}_i$. This method enables the model to focus on each patient's most relevant feature-time pairs, thereby improving personalization and accuracy~\cite{attention_2017}. Additionally, obtaining a separate attention matrix for each patient allows us to distinguish the importance of variables by class.

\subsubsection{Post-HOC}
TimeSHAP is a post-hoc and model-agnostic method designed to explain recurrent models, such as RNNs, used in sequential decision-making tasks~\cite{timeshap2021}. Based on KernelSHAP, TimeSHAP adapts its theoretical foundations to work with sequential data, attributing importance values to features and time steps along a sequence.

SHAP is a game-theoretic explanation technique that assigns importance to the features of an input based on their contribution to the model's output~\cite{shap2017}. KernelSHAP, an implementation of SHAP, approximates Shapley values by sampling coalitions of features and fitting a local linear model~\cite{shap2017}. SHAP focuses on solving an optimization problem that aims to minimize the loss between the original model's output \( f \) and a linear explanatory model \( g \):
\[
f(h_x(\mathbf{z})) \approx g(\mathbf{z}) = w_0 + \sum_{n=1}^{N} w_n z_n,
\]
where \( h_x(\mathbf{z}) \) is the perturbation function that maps binary coalitions \( \mathbf{z} \) to the original input space, $\{ w_n\}_{n=1}^N$ are the feature importance values (Shapley values), and \( w_0 \) is the bias term representing the model output when all features are off. See~\cite{shap2017} for more details on SHAP.

For a sequence \(\mathbf{X} \in \mathbb{R}^{F \times T}\) with \( F \) features and \( T \) time steps, TimeSHAP fits a linear explanatory model to minimize the loss in the sequential context. 
The perturbation function for the two-dimensional matrix is redefined as follows when explaining time steps (columns):
\[
h_{\mathbf{X}}(z) = \mathbf{X} \mathbf{D}_{\mathbf{z}} + \mathbf{B} (\mathbf{I} - \mathbf{D}_{\mathbf{z}}), \quad \mathbf{D}_{\mathbf{z}} = \text{diag}(\mathbf{z}),
\]
where, \(\mathbf{X} \in \mathbb{R}^{F \times T}\) is the original input matrix, \(\mathbf{B} \in \mathbb{R}^{F \times T}\) is a background matrix (typically with the average values of the features across the sequence), and \(\mathbf{D}_{\mathbf{z}} \in \mathbb{R}^{T \times T}\) is a diagonal matrix constructed from the coalition vector \( \mathbf{z} \). This perturbation function allows TimeSHAP to compute the importance of time steps in sequential data, considering the temporal dimension.

The adaptation of TimeSHAP, named IT-SHAP, to handle temporal outputs of models involves computing explanations not only for the final prediction but also for intermediate predictions along the sequence. This is particularly relevant for models that make predictions at each time step, as in TS tasks. For the $i$-th patient \(\mathbf{X}_i \in \mathbb{R}^{F \times T}\), the model produces a sequence of outputs \(\hat{\mathbf{y}} = [\hat{y}_1, \hat{y}_2, \ldots, \hat{y}_T]\). TimeSHAP is adapted to explain each \(\hat{y}_t\) based on all time steps up to time \( t \). The perturbation function is redefined to consider the output at each time step:
\[
h_{\mathbf{X}}(z, t) = [\mathbf{X}]_{(:, 1:t)} \mathbf{D}_{\mathbf{z}_t} + [\mathbf{B}]_{(:, 1:t)} (\mathbf{I} - \mathbf{D}_{\mathbf{z}_t}), \quad \mathbf{D}_{\mathbf{z}_t} = \text{diag}(\mathbf{z}_t),
\]
where \( t \) is the time step for which the perturbation is generated, and \( \mathbf{z}_t \) is the coalition vector at time \( t \). Additionally, \( [\mathbf{X}]_{(:, 1:t)} \in \mathbb{R}^{F \times t} \) and \( [\mathbf{B}]_{(:, 1:t)} \in \mathbb{R}^{F \times t} \). When multiplying \(\mathbf{D}_{\mathbf{z}_t}\) to \(\mathbf{X}\) and \(\mathbf{B}\) on the right, this operation modulates/selects the columns of these matrices that are activated.

To handle irregular MTS, IT-SHAP employs a mask that indicates the presence of time steps up to a maximum \( T \). Although IT-SHAP is applied independently to each patient, it is necessary to adapt it to address the irregularity of MTS, as the model that produces temporal output requires an input of \( T \) time steps and generates an output of \( T \) time steps. Therefore, this mask is needed to adapt the results to the length of each specific patient. Let \(\mathbf{M} \in \{0, 1\}^{F \times T}\) be a binary mask where 
\([\mathbf{M}]_{(f,t)} = 1\) if the time step \( t \) in feature \( f \) exists, and \( 0 \) otherwise. The perturbation function is redefined considering this mask:
\[
h_{\mathbf{X}}(z, t) = ([\mathbf{X}]_{(:, 1:t)} \odot [\mathbf{M}]_{(:, 1:t)}) \mathbf{D}_{\mathbf{z}_t} + [\mathbf{B}]_{(:, 1:t)} (\mathbf{I} - \mathbf{D}_{\mathbf{z}_t}), \quad \mathbf{D}_{\mathbf{z}_t} = \text{diag}(\mathbf{z}_t),
\]
where \(\mathbf{X}_{(:, 1:t)} \odot [\mathbf{M}]_{(:, 1:t)} \in \mathbb{R}^{F \times t}\) represents the masked input \(\mathbf{X}\) up to time \( t \) using the Hadamard product, and \(\mathbf{B} \in \mathbb{R}^{F \times T}\) is the background matrix. Finally, since the diagonal matrix \(\mathbf{D}_{\mathbf{z}_t}\) is multiplying from the right, its effect is to activate/deactivate the columns of matrices $\mathbf{X}_{(:, 1:t)} \odot [\mathbf{M}]_{(:, 1:t)}$ and $[\mathbf{B}]_{(:, 1:t)}$.

IT-SHAP fits a linear model \( g(\mathbf{z}_t) \) to approximate \( f(h_{\mathbf{X}}(\mathbf{z}_t, t)) \), where \( \mathbf{W}_i \in \mathbb{R}^{F \times T} \) indicates feature importance over \( T \) time steps. Here, \([\mathbf{W}_i]_{(f,t)}\) denotes the importance of feature \( f \) at time step \( t \) for the \( i \)-th patient. This provides a detailed explanation of predictions over time. IT-SHAP enhances transparency in recurrent models' decision-making with sequential data, including temporal outputs and irregular TS, fostering trust in critical applications.

\section{Database and Data Preparation}\label{sec:Database}

This study is based on longitudinal clinical data spanning 17 years, acquired from the UHF in Spain, covering the period from January 2004 to February 2020. To safeguard patient confidentiality, a rigorous anonymization protocol was implemented. The analysis focuses on 3,502 patients admitted to the ICU of the UHF during this period, with the endorsement of the UHF Research Ethics Committee under internal reference 24\_22. The study focuses solely on the ICU stay period, excluding pre-admission data, as events before ICU admission are less relevant to patient care and microbial transmission related to MDR.

Pathogens that have developed resistance to multiple antimicrobial agents are known as MDR organisms~\cite{wartu2019multidrug}. The WHO has identified these microorganisms as a significant global health threat due to their limited treatment options and high transmission rates. Recognizing the critical impact of MDR organisms on public health, and underscored by their designation as a top priority by the WHO~\cite{world2024bacterial}, this study aims to investigate the temporal dynamics of various pathogens within the ICU. By analyzing these patterns over time, the study seeks to provide insights that could potentially inform more effective containment and treatment strategies.

To identify MDR pathogens, microbiological cultures followed by antimicrobial susceptibility testing (antibiograms) were conducted, a process that takes at least 48 hours. This study prioritized the initial culture indicating MDR, guided by clinical discretion. Among the 3,502 ICU patients, 548 had at least one MDR culture during their ICU admission, resulting in a significant class imbalance that must be addressed in the analysis. The focus on the ICU stay period, the rigorous identification process for MDR pathogens, and the handling of a significant class imbalance are critical components of this study, ensuring that the analysis is both comprehensive and relevant to current clinical challenges.

\begin{figure}[h]
    \centering
    \includegraphics[width=1\columnwidth]{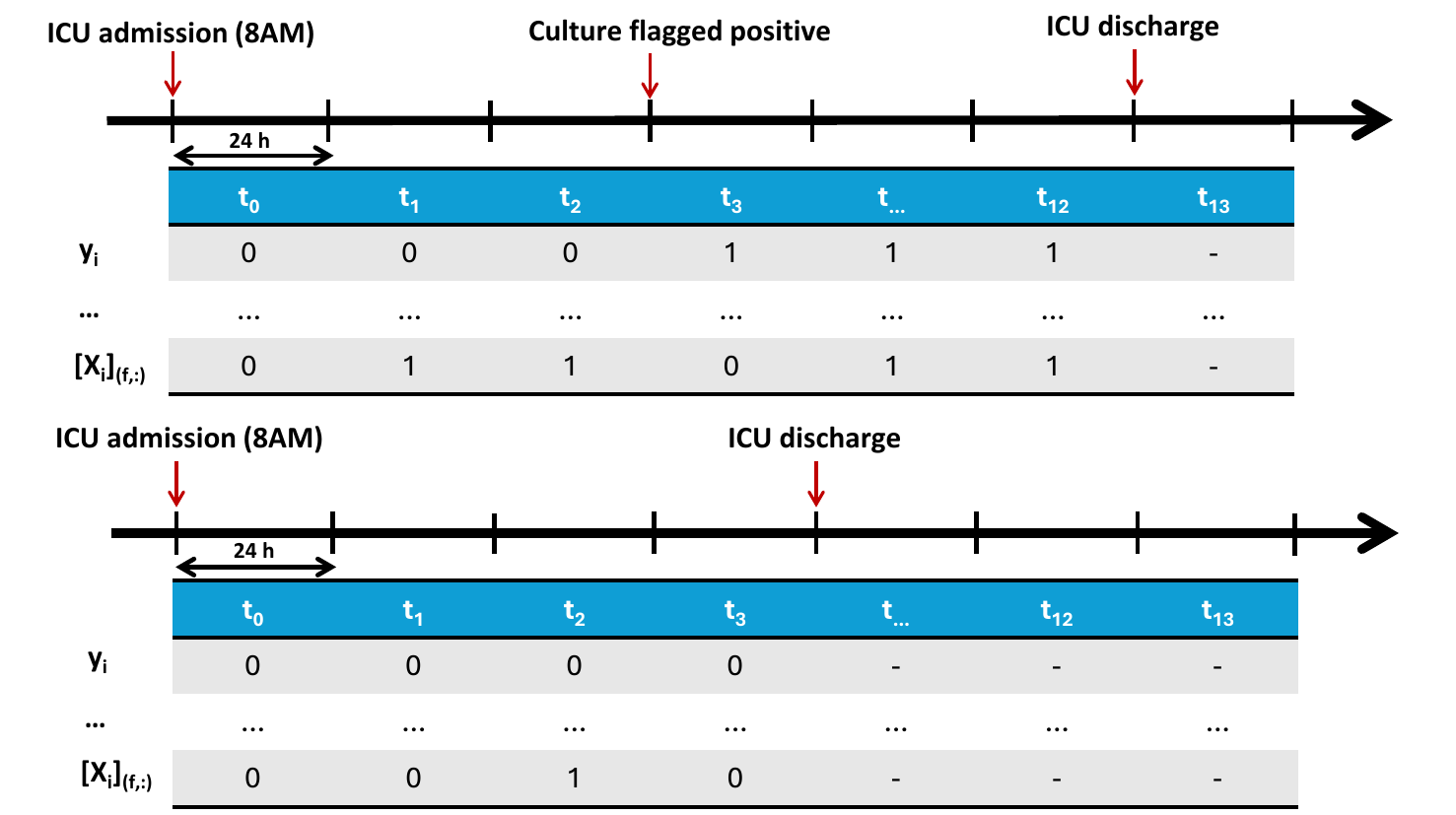}
    \caption{Preprocessing steps for MTS data. The first figure (top) represents the data for an MDR patient, where $\mathbf{y}_i$ indicates the time-varying MDR label, and $\mathbf{X}_{i}$ collects the values of the $F$ features for the $T=14$ time instants. The second figure (bottom) represent the data for a non-MDR patient. The timeline spans from ICU admission (8 AM) to ICU discharge, with a 24-hour sampling period. For MDR patients, if the culture is flagged positive on the day $t_i$, then $[\mathbf{y}_i]_{(t)} = 0$ for $t<t_i$ and $[\mathbf{y}_i]_{(t)} = 1$ for $t\geq t_i$. For non-MDR patients $[\mathbf{y}_i]_{(t)} = 0$ throughout their ICU stay.}
    \label{fig:preprocessing}
\end{figure}

For our MTS processing, the first time step ($t = 0$) is defined as the day at 8 AM when the $i$-th patient is admitted to the ICU. Non-MDR patients are labeled $[\mathbf{y}_{i}]_{(t)} = 0$ for all $t$, while MDR patients have values of $0$ and $1$. From the time instant the culture is identified as MDR, the value is 1 for up to 14 days (see Figure~\ref{fig:preprocessing}). This decision is based on clinical criteria: i) the first two weeks in the ICU are critical for the emergence of MDR germs~\cite{hinman1992meeting}, and ii) UHF clinical protocols quarantine MDR patients for 14 days~\cite{thombley2010menu}. A 24-hour sampling period, from 8 AM to 8 AM the next day, was used, aligned with the ICU work organization. Studies have shown that models using relatively long windows with MTS of irregular length perform better in predicting MDR occurrence than those using shorter windows or imputing missing values~\cite{martinez2022interpretable}. 

Each patient's MTS includes four primary groups of information: culture-related features, antibiotic treatments, ICU occupancy and co-patient treatments, and individual patient care. Co-patients are those who remain in the ICU during the same time intervals as the patient under study, excluding the patient in question.

Culture-related variables are used to identify time intervals during which specific germs are detected. We track six germs known for their potential MDR: \textit{Pseudomonas}, \textit{Stenotrophomonas}, \textit{Acinetobacter}, \textit{Enterobacter}, \textit{Staphylococcus aureus}, and \textit{Enterococcus}. These binary variables are denoted by the germ name followed by the subscript \textsubscript{pc} (previous cultures). An additional variable, \textit{Others}\textsubscript{pc}, indicates whether the $i$-th patient had positive germs but was not resistant, not included in the aforementioned six groups, in previous cultures.
This variable provides context on potential precursors to resistant strains. 
It is important to note that, as we aim to predict the first MDR infection in each patient, the germs modeled in the six previous culture variables were not resistant; they merely provided information on previous cultures. By clinical definition, \textit{Stenotrophomonas} (specifically \textit{Stenotrophomonas maltophilia}) are always multiresistant, as are enterobacteria Extended-Spectrum Beta-Lactamase (ESBL). Consequently, we do not have non-multiresistant previous cultures for \textit{Stenotrophomonas}, and for \textit{Enterobacter}, we only include those that are not ESBL.

Antibiotic treatment features include clinically relevant antibiotic families: Aminoglycosides (AMG), Carbapenems (CAR), 1\textsuperscript{st} generation Cephalosporins (CF1), 3\textsuperscript{rd} generation Cephalosporins (CF3), 4\textsuperscript{th} generation Cephalosporins (CF4), unclassified antibiotics (\textit{Others}), Glycopeptides (GLI), Lincosamides (LIN), Lipopeptides (LIP), Macrolides (MAC), Nitroimidazoles (NTI), Oxazolidinones (OXA), Broad-spectrum Penicillins (PAP), Penicillins (PEN), Polypeptides (POL), Quinolones (QUI), and Sulfonamides (SUL). The ``Others" feature identifies any additional antibiotic families not listed above. Each variable is binary, indicating whether the corresponding antibiotic has been administered or not

Concerning the third group of features, we include both the occupancy of the ICU and a summary of the antimicrobials administered to the other patients in the ICU (referred to as ``neighbors'' or ``co-patients'') during the same time intervals considered for the patient under study. Thus, we modeled 20 additional numerical features: the number of neighbors  (\# of pat\textsubscript{tot}), the total number of patients receiving any antibiotics (\# of pat\textsubscript{atb}), the number of patients identified with MDR bacteria (\# of pat\textsubscript{MDR}), and the number of neighbors receiving each of the 17 antibiotic families mentioned above, that coexist with the patient in question. These numerical time features for each patient are referred to as ``environmental variables", which act as indicators of the health status of co-patients in the ICU. The incorporation of these ICU variables, a fundamental aspect of our approach, is well-supported by clinical knowledge and has demonstrated significant benefits in previous studies focused on MDR classification. We use the subscript \textsubscript{n} in the variable name to denote characteristics that refer to the patient's neighbors under study. 

Lastly, variables related to individual patient care in the ICU are considered, categorized into continuous and binary types.
Continuous variables encompass the duration of mechanical ventilation, number of transfusions, incidences and durations of tracheostomy, and usage durations of ulcers and hemofilters. These variables offer crucial insights into the patient's clinical status and treatment protocols. Binary variables include whether a patient underwent more than three postural changes, a preventive measure against immobility-related complications in ICU settings. Also included is whether insulin was administered for over six hours within a specific interval, whether daily artificial nutrition exceeded six hours (essential for critically ill patients unable to feed orally and relevant in MDR development), and whether sedation and relaxation each exceeded six hours daily, evaluated individually for each patient at specific time points. Furthermore, organ failures were assessed to identify instances where patients experienced more than two specific failures—such as hepatic, renal, coagulation, hemodynamic, or respiratory failures—at a given time point. This metric provides critical insights into the severity of each patient's condition and may necessitate additional medical interventions. Organ failures are referenced using the subscript ``fail" in subsequent sections to denote these characteristics. Additionally, the administration of vasoactive drugs was documented.

\section{Experiments and Results}
\label{sec:DataAnalytic}
This section begins with a brief description of the experimental setup. Subsequently, we conduct temporal prediction of MDR patients, proposing an integrated approach that employs a GRU along with various interpretability methods. Following this, we explore the explainability provided by the proposed methods. Finally, we perform a comparative analysis of the performance in terms of prediction and explainability offered by the different methods.

\subsection{Experimental Setup}
\label{sec:ExperimentalSetup}

The dataset $\mathcal{D}$ was randomly divided into two independent subsets: 70\% for training and 30\% for testing. The training set was used to build the model, while the test set was used to evaluate its performance. 
Considering class imbalance, model performance was assessed using sensitivity, specificity, and Receiver Operating Characteristic Area Under the Curve (ROC AUC) metrics~\cite{roc_2022, imbalance_2022}. Sensitivity identifies true positives (MDR cases), ensuring proper management of at-risk patients. Specificity identifies true negatives (non-MDR cases), reducing false positives and unnecessary interventions. ROC AUC provides an aggregate performance measure across all classification thresholds. To prevent bias from a single random split, each experiment was repeated three times with different train-test partitions.

As noted in Section~\ref{sec:Database}, the dataset exhibits binary class imbalance, potentially biasing the model towards the majority class~\cite{imbalance_2022}. To mitigate this, we employed 5-fold cross-validation during training to identify hyperparameters that minimize the Temporal Balanced Binary Cross Entropy (TBBCE) cost function. This function, a modification of the binary cross-entropy, incorporates a weight $\beta_t \in (0,1)$ for each time step $t$ to adjust the penalty, ensuring adequate representation of the minority class~\cite{aurelio2019learning}. Formally, the TBBCE cost function is defined as:

\begin{equation}
    \mathcal{L}_{\text{TBBCE}} = \frac{-1}{I'T} \! \sum_{i=1}^{I'} \!\sum_{t=1}^{T} \!\left( \beta_t [\mathbf{y}_i]_{(t)} \log([\hat{\mathbf{y}}_i]_{(t)}) + (1 \!-\! \beta_t)(1\! -\! [\mathbf{y}_i]_{(t)}) \log(1 \!- \![\hat{\mathbf{y}}_i]_{(t)}) \right)\!,
\end{equation}

\noindent where $I'$ is the total number of samples in the training set, and $T$ is the total number of time steps. Following~\cite{aurelio2019learning} and others previous research~\cite{martinez2022interpretable, martinez2023LSTM}, the weight $\beta_t$ is set as the ratio of majority class samples to the total number of samples.

To prevent overfitting, an early-stopping strategy was employed, monitoring validation loss at each training epoch~\cite{early_stop_2022}. The hyperparameters tuned for the GRU architecture included: (i) dropout rate, (ii) learning rate, and (iii) number of neurons in the hidden layer.

\subsection{Prediction Results}

\begin{figure}[h]
\centering
	\begin{subfigure}[]
		\centering
        \includegraphics[width=0.475\textwidth]{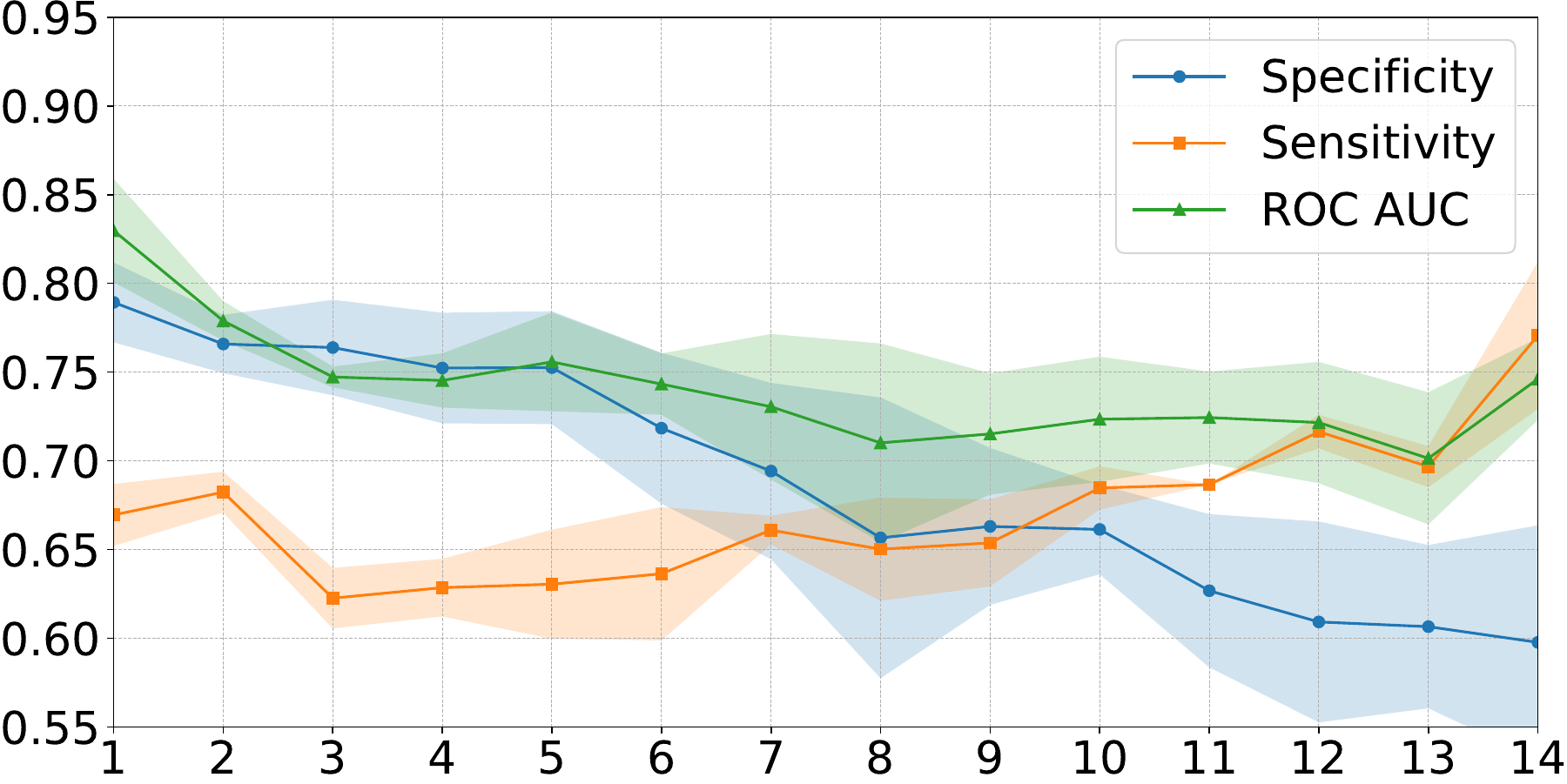}
	\end{subfigure}
	\begin{subfigure}[]
		\centering
        \includegraphics[width=0.475\textwidth]{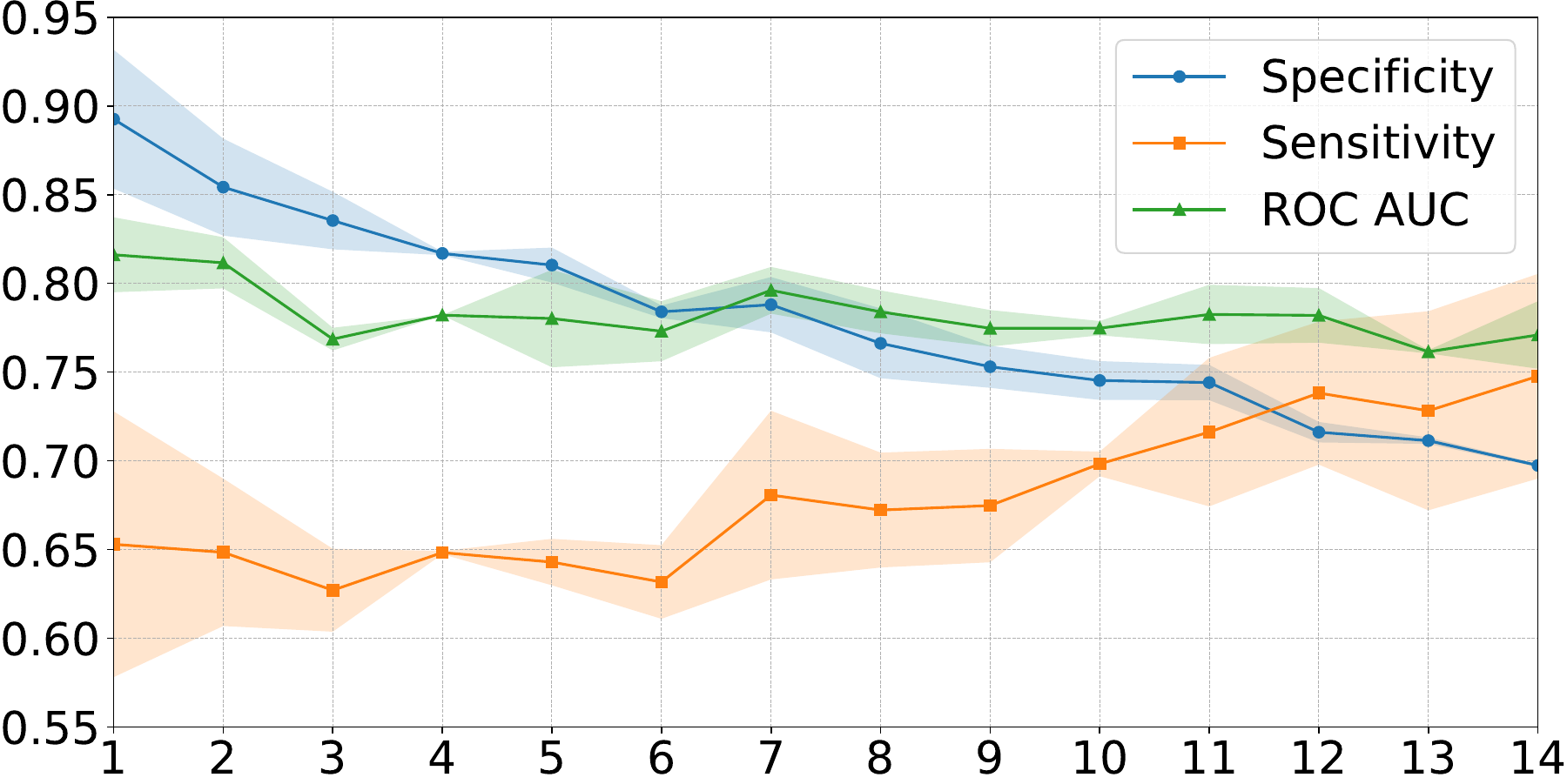}
	\end{subfigure}

    \caption{Performance of the predictive model over time. (a) GRU model performance, and (b) GRU model with an attention mechanism. Each plot displays the mean and standard deviation of specificity, sensitivity, and ROC AUC across different time points.}
    \label{performance}
\end{figure}

This section presents the results obtained from the proposed methodology for classifying patients as MDR or non-MDR over time. Figure~\ref{performance} shows the performance of the predictive model, where Figure~\ref{performance}~(a) displays the results for the GRU model and Figure~\ref{performance}~(b) shows the results for the GRU model enhanced with an attention mechanism. Both panels visualize three metrics: specificity, sensitivity, and ROC AUC, allowing for a comprehensive analysis of the model's classification capabilities over time.

Figure~\ref{performance}~(a) illustrates the mean and standard deviation for each time point. Initially, the specificity is high but decreases as patients remain longer in the ICU. 
This clinically coherent trend occurs because prolonged ICU stays often correlate with deteriorating health conditions and the development of nosocomial infections. These infections can become MDR, posing challenges for accurately classifying patients as MDR or non-MDR in later time steps~\cite{trubiano2015nosocomial}.
Conversely, sensitivity increases over time, suggesting that longer ICU stays provide more patient information, enabling the model to better identify MDR patients. On the other hand, Figure~\ref{performance}~(b) presents the results for the GRU model with Hadamard attention. 
While the specificity follows a trend similar to the one observed for the basic GRU model, the absolute values are better and the relative decay is slightly smaller. Moreover, the sensitivity provides more stable results. As a result, the ROC improves and exhibits a more stable behavior across time, indicating that the attention mechanism helps maintain classification performance even during extended ICU stays.

The model's overall performance, as verified by clinical experts, represents a significant advancement in addressing MDR. The GRU model can classify whether a patient will develop MDR with an average sensitivity of 67.07\% and an average specificity of 68.98\%, achieving an average ROC AUC of 74.08\%. In contrast, the GRU model enhanced with attention mechanisms improves classification performance, achieving an average sensitivity of 67.91\% and an average specificity of 77.95\%, with an average ROC AUC of 78.27\%. These insights facilitate prompt interventions, such as preventive isolation upon patient admission. To put these results into perspective, a non-time-varying model that attempts to address the MDR issue, along with a non-recurrent neural network used to classify patients as either MDR or non-MDR, yields an ROC AUC of 70.35\%~\cite{rw_xai3_2023_transAMR}. Additionally, the previous state-of-the-art results for an earlier version of this database achieved an ROC AUC of 66.73\%~\cite{martinez2022interpretable}. Our enhanced model, therefore, not only surpasses these benchmarks but also offers substantial improvements in clinical decision-making and patient management.

\subsection{Interpretability Results}

After analyzing the classification results, we now focus on interpretability. 
We employed three different methods to achieve interpretability (see Section~\ref{sub:xai}), each aiming to identify the most important variables and potential risk factors that provide greater clinical knowledge for addressing MDR acquisition in the ICU.

It is necessary to define the meaning of the importance ranges for each method. CMI and Hadamard represent the importance of variables on a scale from 0 to 1, where 0 indicates no importance and 1 indicates maximum importance. Conversely, IT-SHAP provides both positive values (indicating contribution to the class in question) and negative values (indicating irrelevance or contribution to the non-MDR class). CMI does not require an ML/DL model, Hadamard is an intrinsic method within the GRU, and IT-SHAP is a post-hoc method (for more details see Section~\ref{sub:xai}).

\begin{figure}[h!]
\centering
	\begin{subfigure}[]
		\centering
        \includegraphics[width=0.32\textwidth]{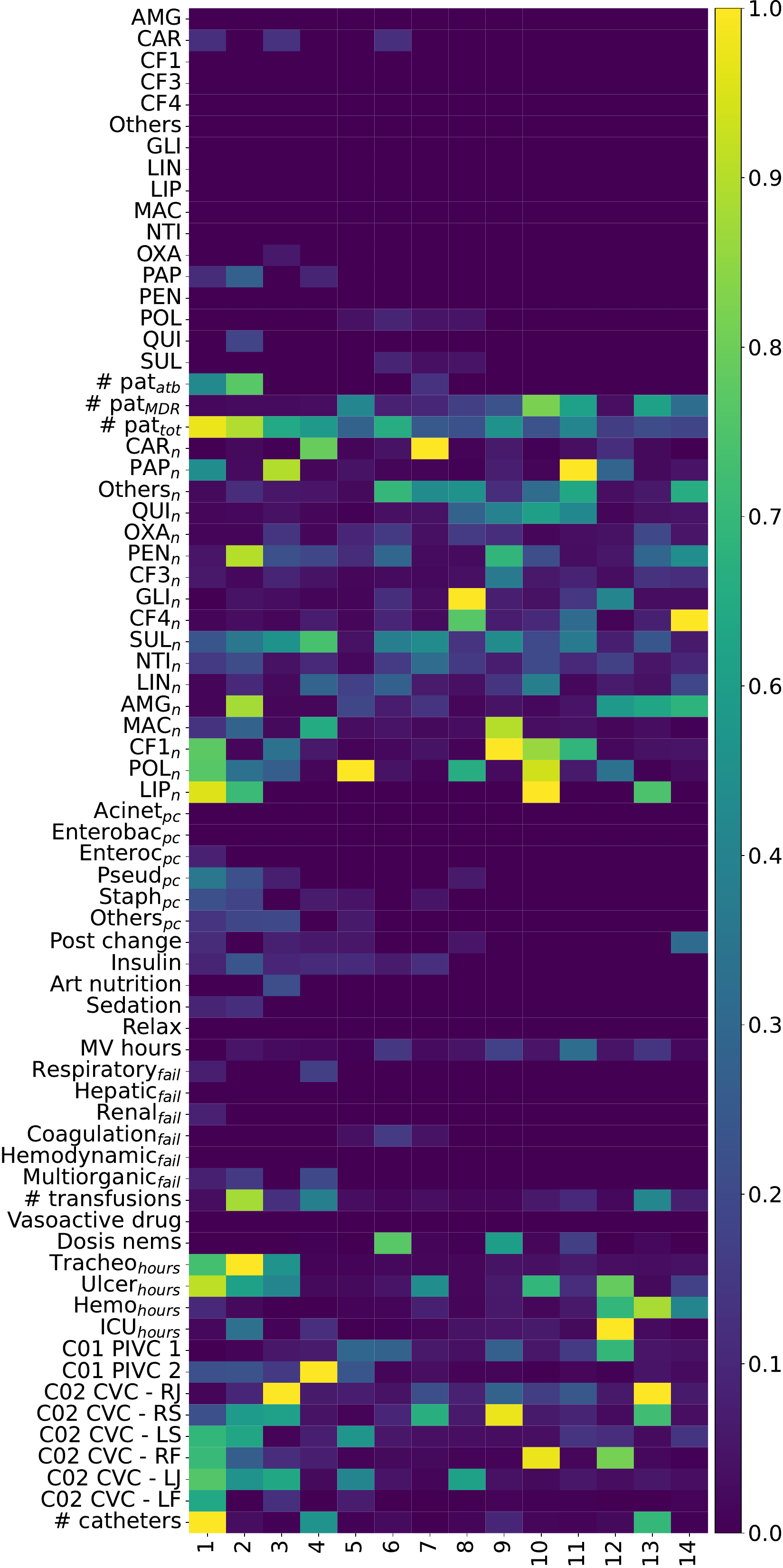}
	\end{subfigure}
	\begin{subfigure}[]
		\centering
        \includegraphics[width=0.32\textwidth]{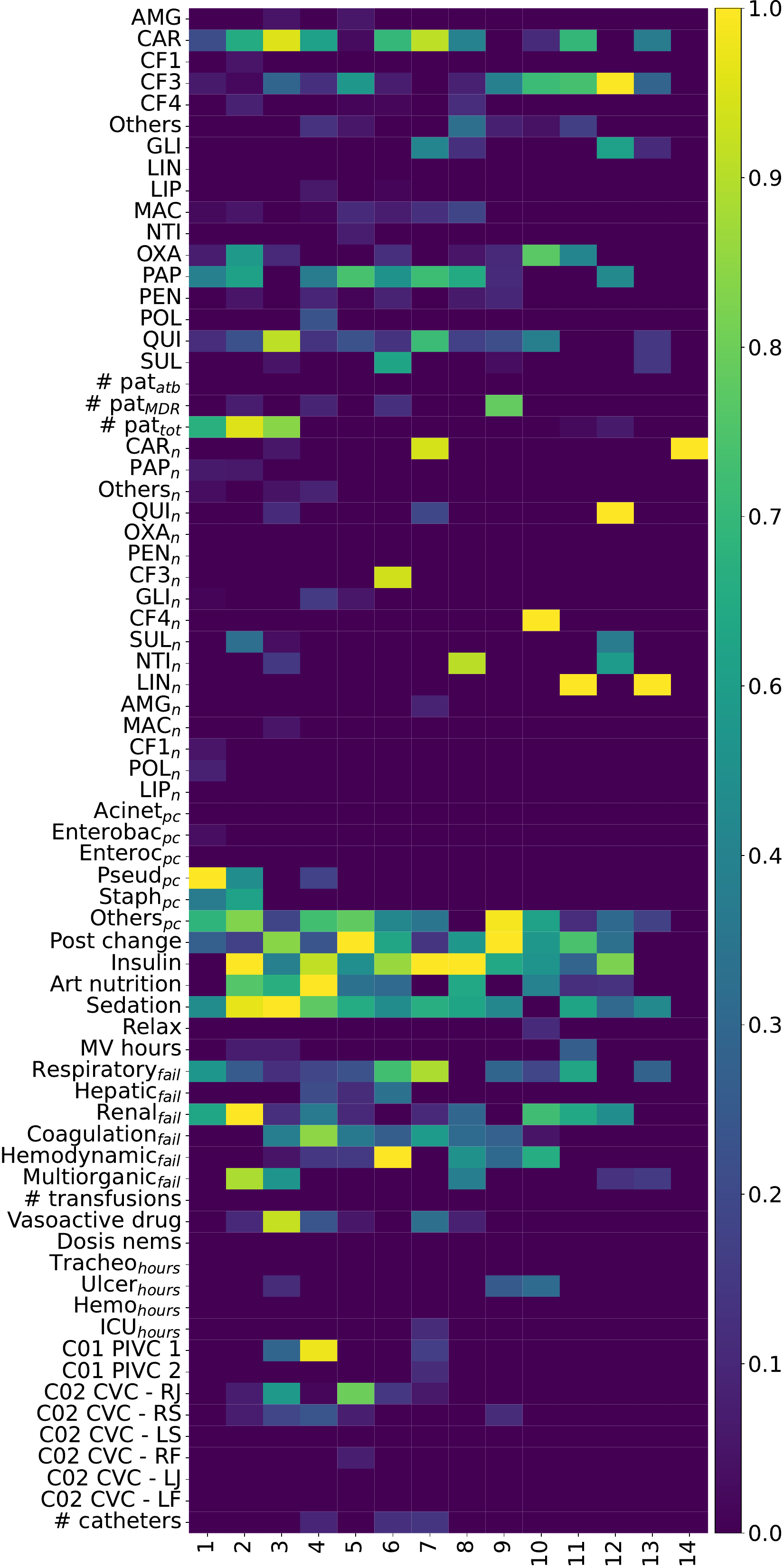}
	\end{subfigure}
    \begin{subfigure}[]
		\centering
        \includegraphics[width=0.32\textwidth]{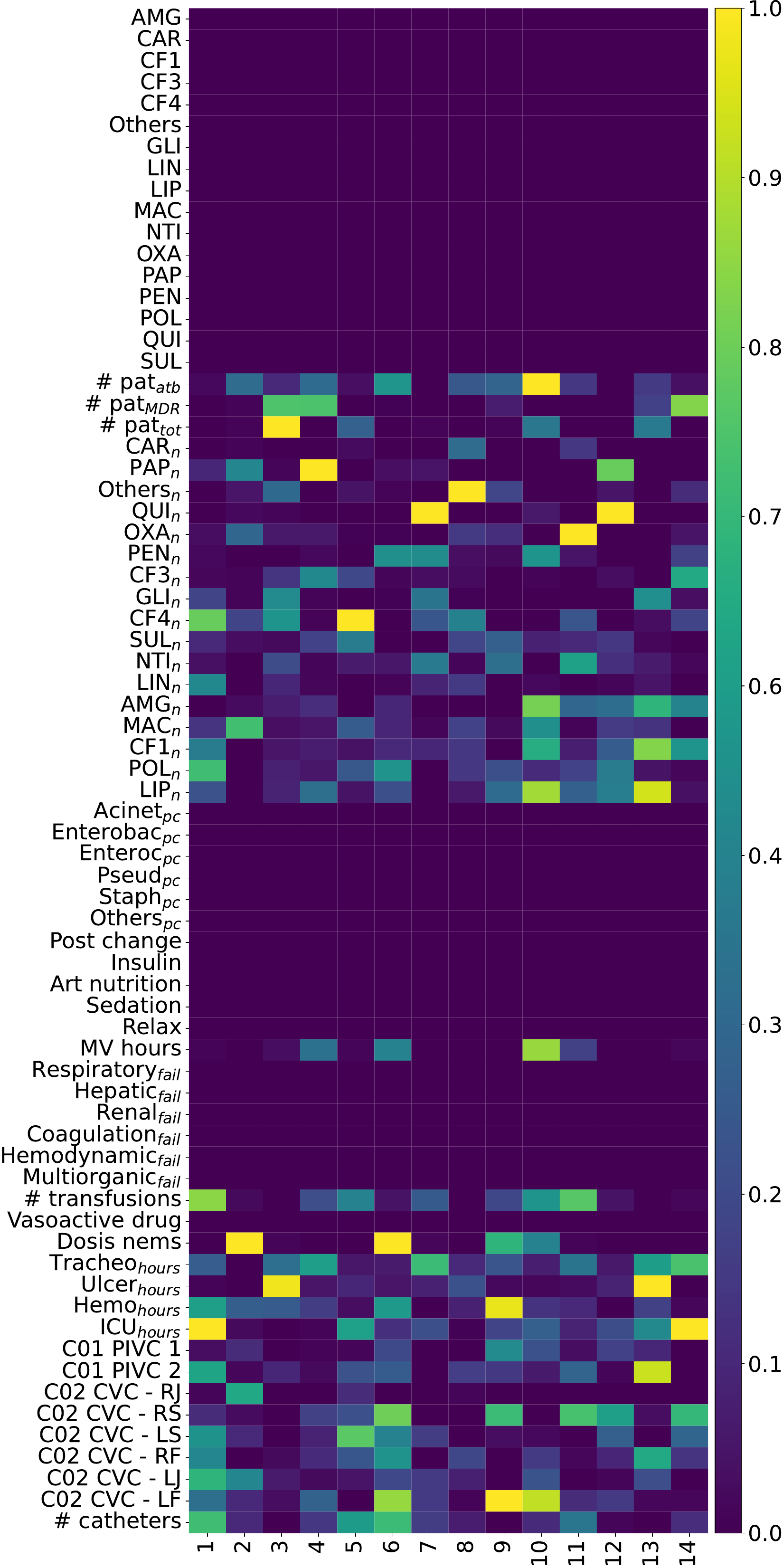}
	\end{subfigure}
    \caption{Pre-hoc interpretability using CMI. (a) Importance scores for the entire population; (b) Importance scores for MDR patients; and (c) Importance scores for non-MDR patients. The x-axis represents the time steps, and the y-axis represents the variables. The color scale indicates the importance of each variable at each time step.}
    \label{cmi}
\end{figure}

\begin{figure}[h!]
\centering
	\begin{subfigure}[]
		\centering
        \includegraphics[width=0.315\textwidth]{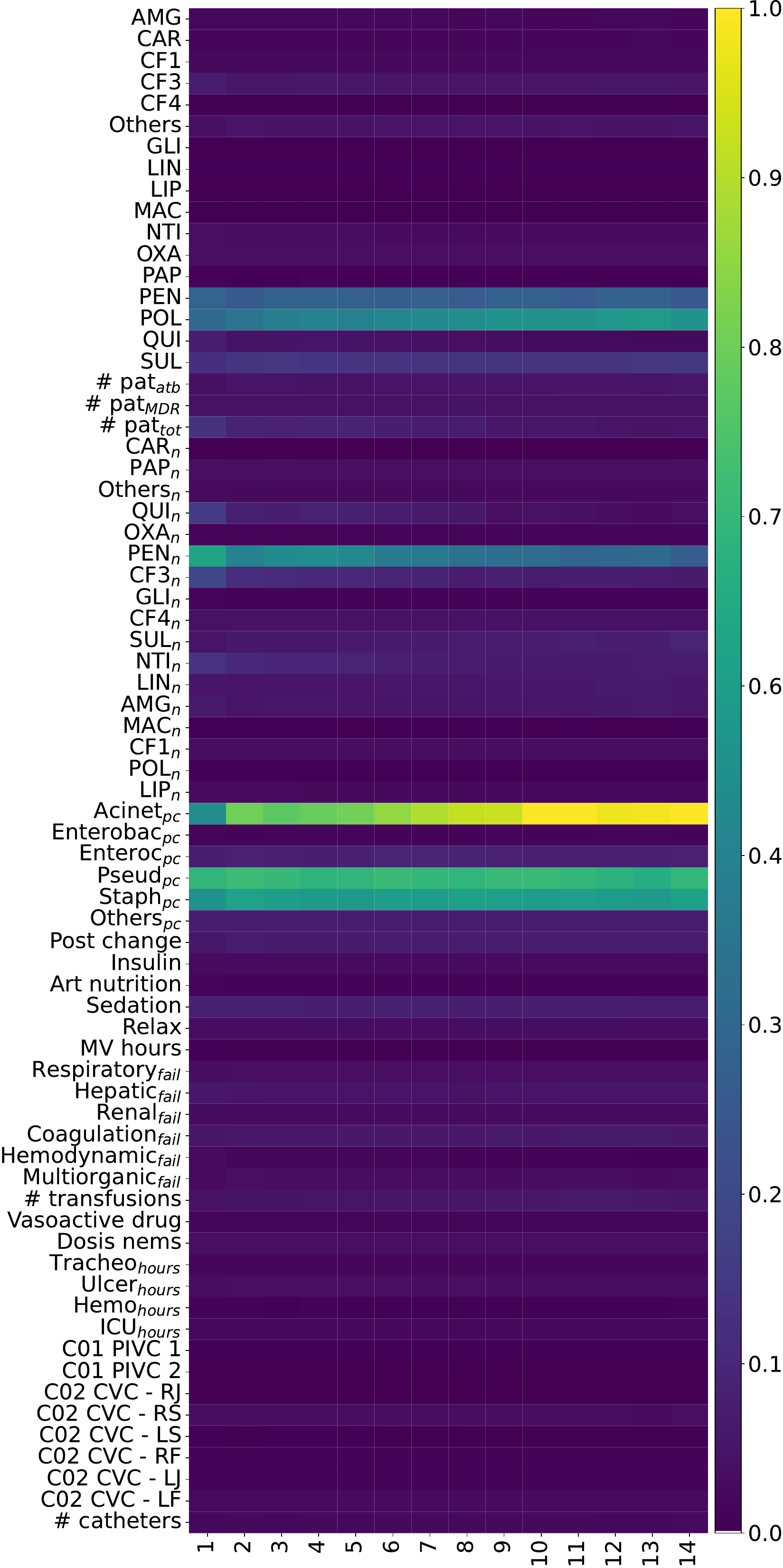}
	\end{subfigure}
	\begin{subfigure}[]
		\centering
        \includegraphics[width=0.315\textwidth]{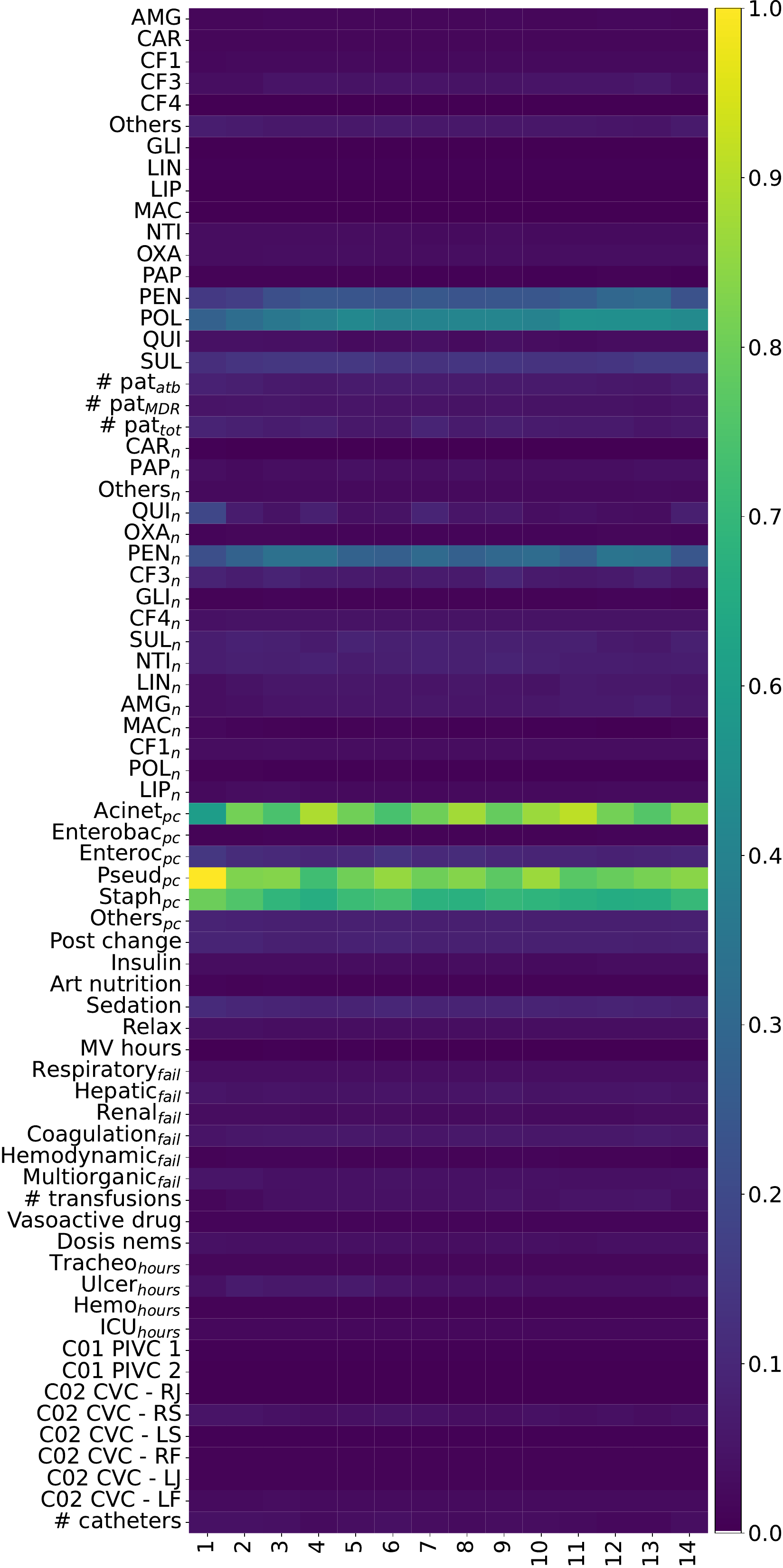}
	\end{subfigure}
    \begin{subfigure}[]
		\centering
        \includegraphics[width=0.315\textwidth]{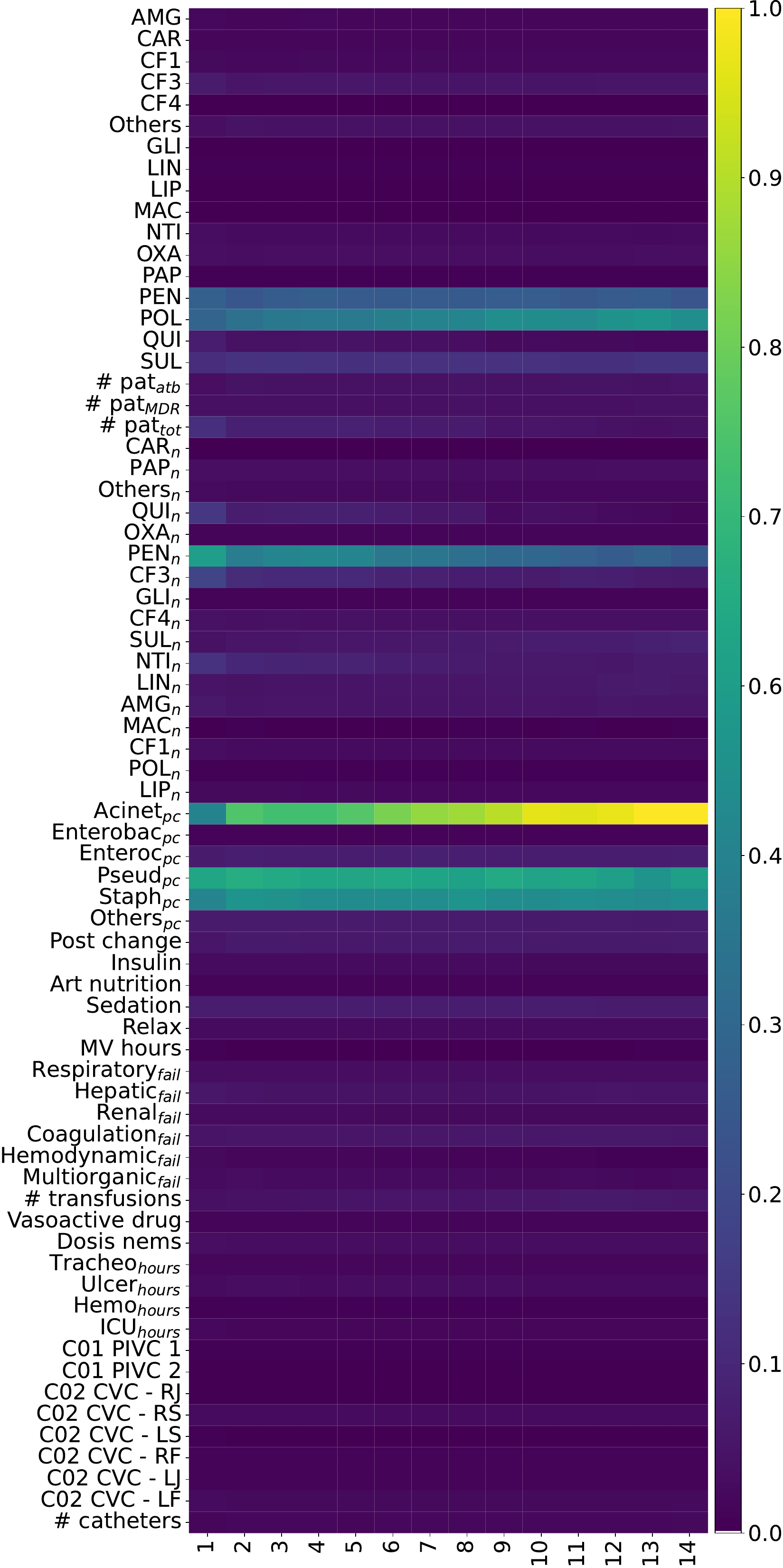}
	\end{subfigure}

    \caption{Intrinsic interpretability using Hadamard attention mechanism. (a) Attention weights for the entire population; (b) Attention weights for MDR patients; and (c) Attention weights for non-MDR patients. The x-axis represents the time steps, and the y-axis represents the variables. The color scale indicates the importance of each variable at each time step.}
	\label{hadamard}
\end{figure}

\begin{figure}[h!]
\centering
	\begin{subfigure}[]
		\centering
        \includegraphics[width=0.315\textwidth]{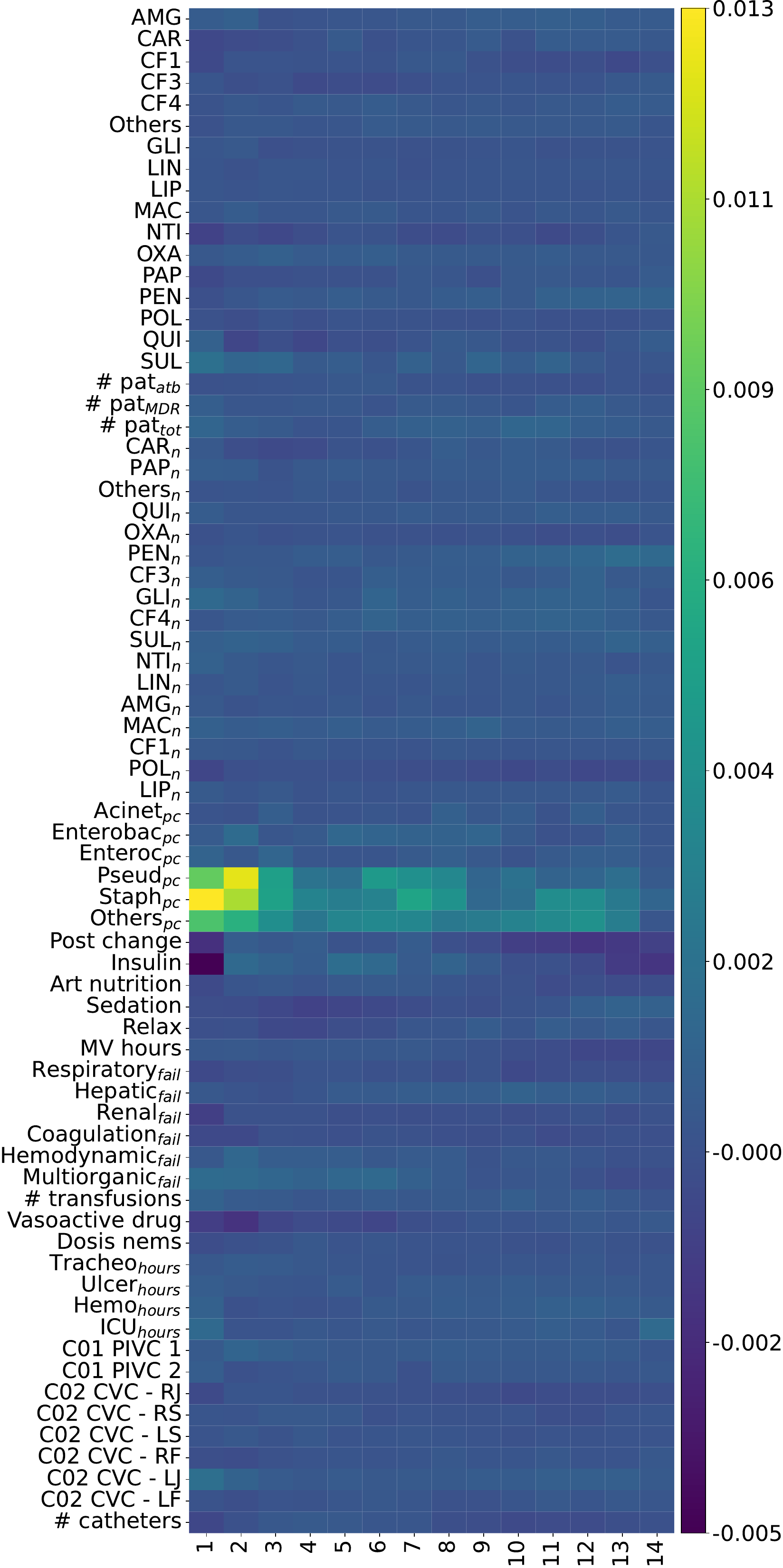}
	\end{subfigure}
	\begin{subfigure}[]
		\centering
        \includegraphics[width=0.315\textwidth]{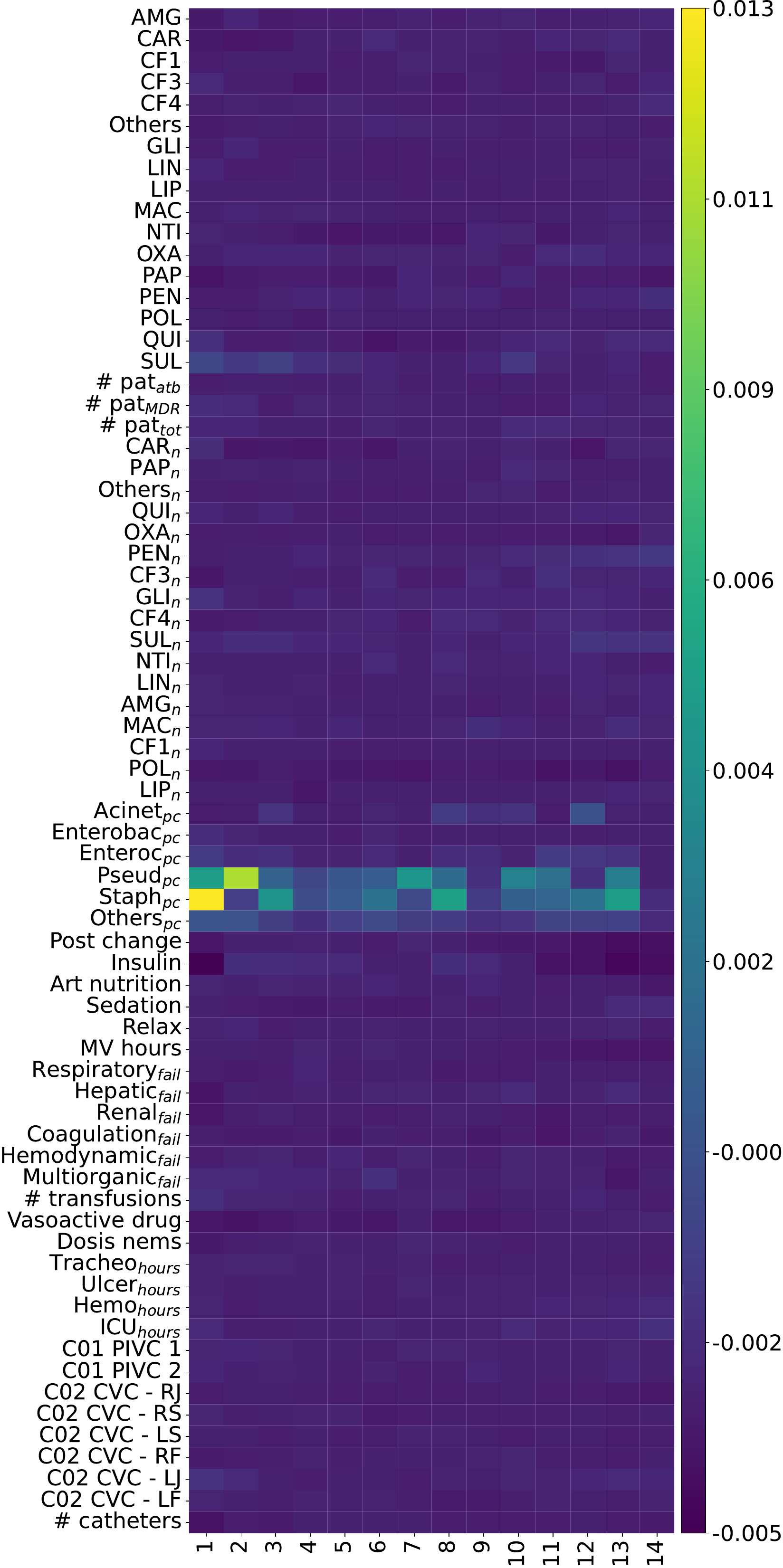}
	\end{subfigure}
    \begin{subfigure}[]
		\centering
        \includegraphics[width=0.315\textwidth]{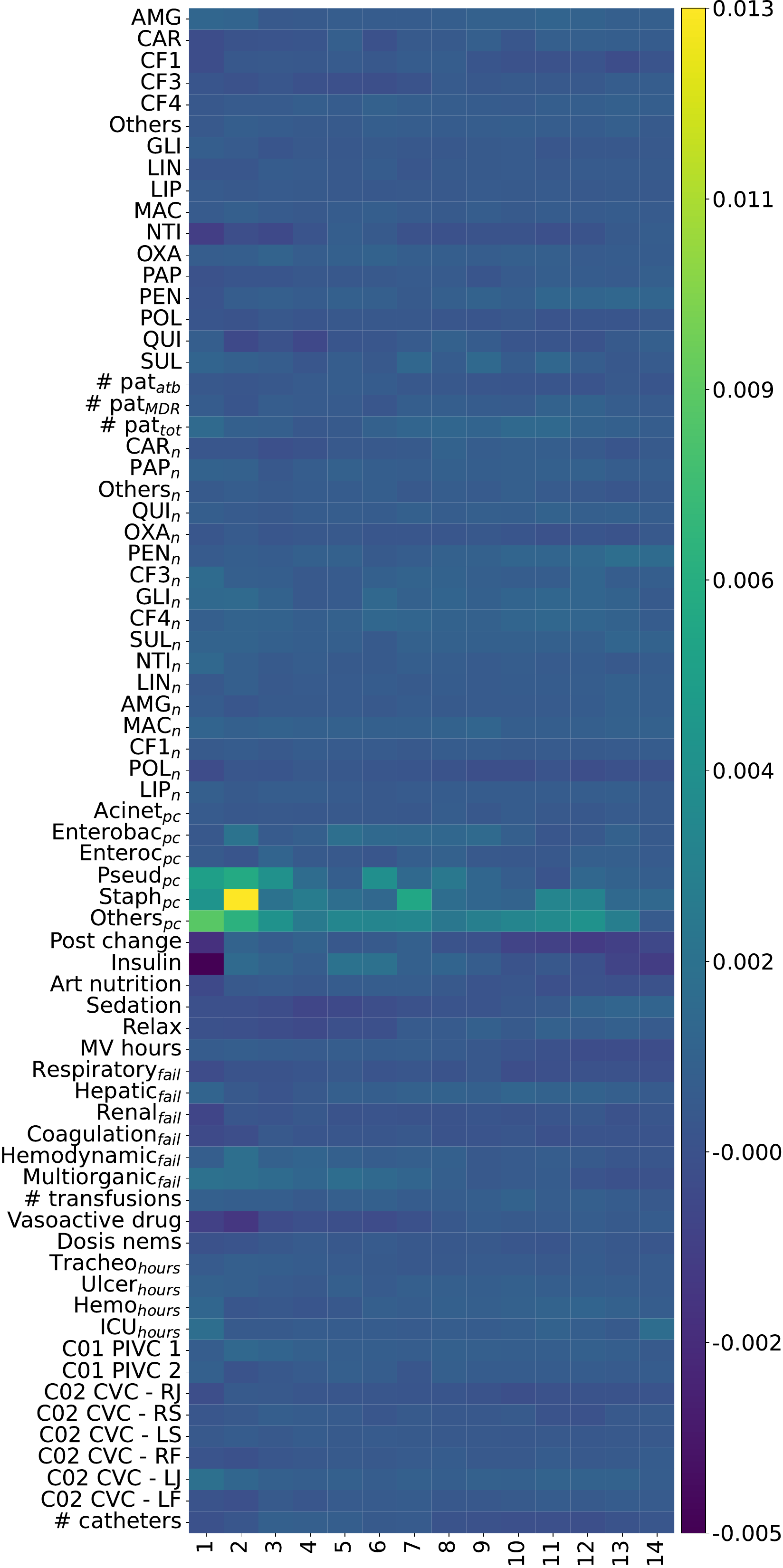}
	\end{subfigure}
    \caption{Post-hoc interpretability using IT-SHAP. (a) Importance scores for the entire population; (b) Importance scores for MDR patients; and (c) Importance scores for non-MDR patients. The x-axis represents the time steps, and the y-axis represents the variables. The color scale indicates the importance of each variable at each time step.}
    \label{shap}
\end{figure}

We begin by explaining the common structure for analyzing the importance of the XAI methods considered (see Figures~\ref{cmi},~\ref{hadamard}, and \ref{shap}). The x-axis of these figures represents the time steps, ranging from 1 to 14, with 1 corresponding to ICU admission. The y-axis displays the variables analyzed, as described in Section~\ref{sec:Database}. Each feature-time step tuple shows the importance of feature \( f \) at time step \( t \). These figures consist of three charts, each providing a different perspective on variable importance: (a) overall population analysis (MDR and non-MDR patients), (b) MDR patient analysis, and (c) non-MDR patient analysis.

The first method analyzed is CMI, illustrated in Figure~\ref{cmi}. Figure~\ref{cmi}~(a) reveals that drawing general clinical conclusions is more challenging. Nonetheless, certain variables such as antibiotic treatment, presence of positive cultures in various germs (labeled as \textit{Others}\textsubscript{pc}), treatments like insulin, artificial nutrition, and sedation, as well as information on failures, appear to have greater importance according to this method. In Figure~\ref{cmi}~(b), which corresponds to the prediction of MDR, ICU occupancy, co-patient treatment, and individual patient care variables—such as hours with tracheotomy, ulcer presence, hemodialysis, and the hours and types of catheters—seem to indicate a higher severity of the patient's condition. This increased severity is correlated with the development of MDR. Conversely, Figure~\ref{cmi}~(c) shows that for non-MDR patients, environmental factors influence the development of germs, but these germs are not MDR. Additionally, while invasive methods aggravate the patient's condition, they do not lead to the development of MDR. Despite providing more specific insights into the variables, this method is highly sensitive to noise and parameter selection, which can affect the conclusions' reliability.

Next, we analyze the Hadamard method, illustrated in Figure~\ref{hadamard}. In Figure~\ref{hadamard}~(a), various groups of more relevant variables are observed: (i) previous cultures associated with \textit{Acinetobacter}, \textit{Pseudomonas}, and \textit{Staphylococcus}; (ii) antibiotics taken by patients, such as PEN and POL; and (iii) co-patient information, including QUI, PEN, and CF3. The importance of these variables varies over time, while the rest of the variables have residual or no importance. On the other hand, Figure~\ref{hadamard}~(b) shows that the presence of positive but non-multiresistant cultures of \textit{Staphylococcus}, \textit{Pseudomonas}, and \textit{Acinetobacter} is highly important in the initial time steps, maintaining importance over time. This clinically suggests that patients admitted to the ICU with a non-multiresistant positive culture in these bacteria will eventually develop MDR germs. Additionally, the influence of antibiotics such as PEN, POL, and to a lesser extent SUL, as well as the influence of the environment, where a neighbor taking PEN could favor the development of MDR, is highlighted. For non-MDR patients (Figure~\ref{hadamard}~(c)), the model focuses again on the presence of cultures of the three previously mentioned bacteria, although in this case with less weight for \textit{Staphylococcus} and \textit{Pseudomonas}, and with great importance for \textit{Acinetobacter} in the last time steps. Also, in the initial time step, there is a slight presence of neighbors taking QUI, PEN (which remains over time), CF3, and NTI, suggesting that the environment affects the development of germs in the patient, but these will not be MDR and will maintain their susceptibility state over time. The presence of a positive \textit{Acinetobacter} culture in the last time steps would indicate that such an infection should not pose an MDR risk.

Finally, we consider the IT-SHAP method, as shown in Figure~\ref{shap}. Figure~\ref{shap}~(a) illustrates that, for both classes, the presence of non-resistant positive cultures plays a crucial role in classification over time. This suggests that prior knowledge of these cultures can significantly influence the prediction of MDR and non-MDR classes. Figure~\ref{shap}~(b) reveals that the presence of positive cultures of \textit{Staphylococcus} and \textit{Pseudomonas} in the initial time steps suggests that the patient is likely to develop MDR germs. Additionally, slight importance of the antibiotic SUL in the initial time steps is observed, although no other significant variables contributing to the model's predictions are identified. It is important to highlight that the administration of insulin and postural changes do not appear to contribute to determining whether the patient will develop MDR. Finally, in Figure~\ref{shap}~(c), it is observed that the presence of previously positive cultures in different germs (\textit{Others}\textsubscript{pc}) allows the model to determine that the patient will not develop MDR over time. This indicates that identifying a greater variety of non-dangerous germs may signal a lower risk of developing MDR. Additionally, information about previous cultures of \textit{Staphylococcus} and \textit{Pseudomonas}, as well as distinct types of failures such as hepatic and multiorgan failure, are relevant for classifying a patient as non-MDR.

All interpretability conclusions have been supported and developed by a specialized clinician. For the specific problem addressed in this study, the Hadamard method provides interpretability results that are more aligned with clinical knowledge, offering precise details about important variables. Although IT-SHAP offers rigorous interpretability consistent with clinical understanding, Hadamard provides a superior level of detail, making it particularly valuable in this context. On the other hand, CMI offers the most dispersed results in terms of reliable knowledge extraction and risk factor identification. Despite yielding more specific insights into variables, CMI sensitivity to noise and parameter selection can affect the reliability of its conclusions. This analysis has been conducted for the entire population and each patient class (MDR and non-MDR), identifying the most relevant variables for each group. This detailed classification allows for more precise identification of risk factors and critical variables contributing to the development or prevention of MDR in ICU patients.

\subsection{Performance and Interpretability Analysis}

\begin{figure}[h!]
    \centering
	\centering
	\includegraphics[width=0.6\columnwidth]{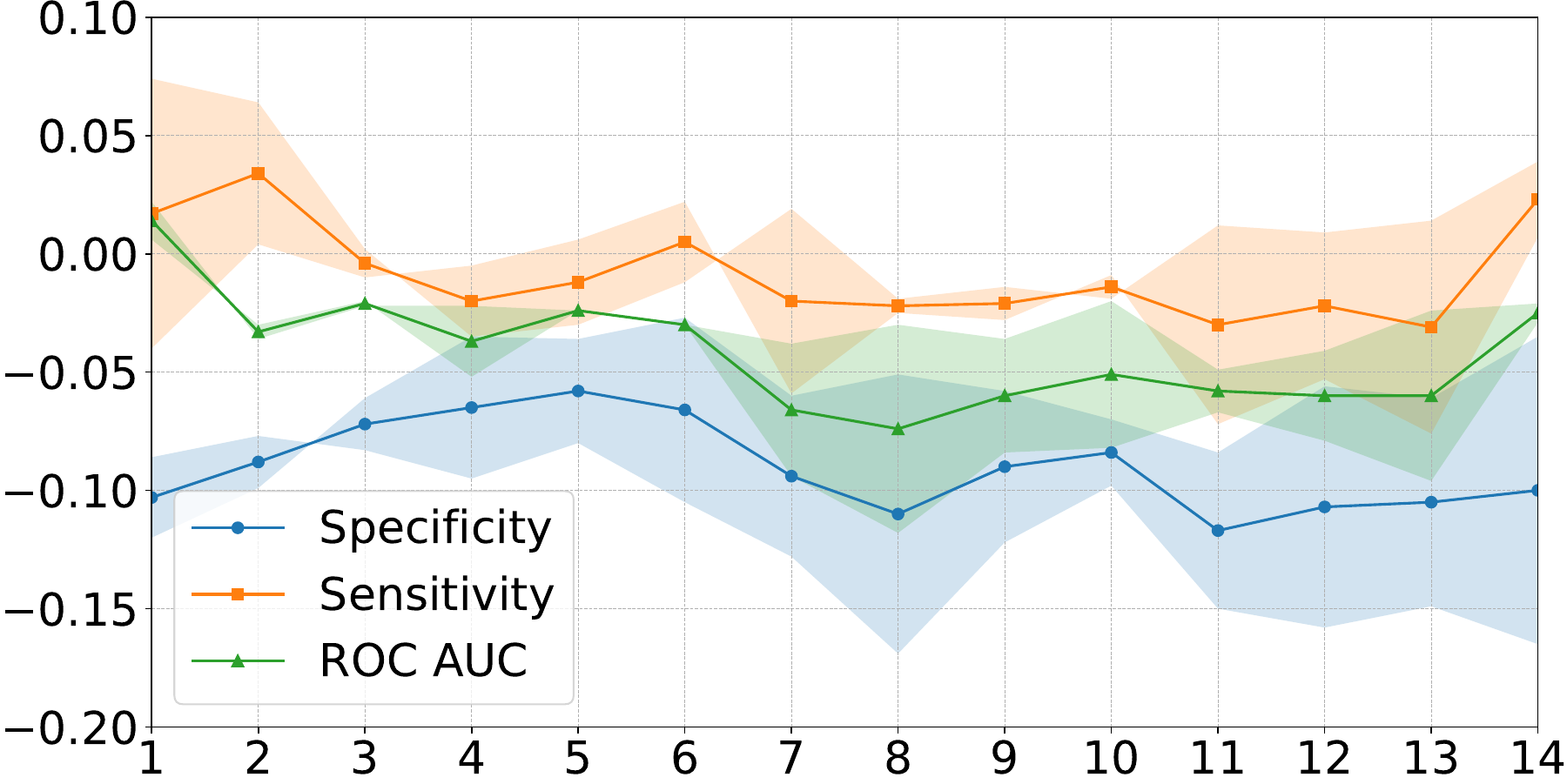}
    \caption{Graphical illustration of the performance comparison between the original GRU model and the GRU model with an attention mechanism. The difference in mean and standard deviation between models is shown. Negative values indicate that the GRU model with an attention mechanism outperforms the original GRU model.}
    \label{comparison}
\end{figure}

Subsequently, we analyzed the comparison between predictive performance and knowledge extraction (interpretability) among the different XAI methods proposed in this work. Figure~\ref{comparison} illustrates the difference in mean and standard deviation for each time step between the original GRU model and the GRU model with an attention mechanism. The results demonstrate that the GRU model with attention improves specificity values at all time steps and nearly improves ROC AUC and sensitivity in most time steps compared to the original GRU model. Therefore, our models do not suffer significantly from the performance-interpretability trade-off reported in the literature~\cite{tradeoff_xai_2020}. The intrinsic attention method, which might be expected to have worse metrics, has been able to match and even surpass the original model in most time steps.

From a clinical perspective, the interpretability provided by CMI generates more dispersed knowledge regarding existing clinical knowledge, while Hadamard and IT-SHAP offer more coherent and conservative insights. However, CMI is not associated with any prediction model and is extremely sensitive to noise and parameter selection~\cite{kraskov2004estimating}. 
The methodology proposed in this work allows us to extract the following insights, identifying risk factors and conditions that can help clinicians in their decision-making when admitting similar patients:

\begin{itemize}
    \item If a patient presents, upon ICU admission, any non-resistant positive culture in dangerous bacteria such as \textit{Staphylococcus} and \textit{Pseudomonas}, the mechanism of action should involve implementing preventive isolation to prevent possible cross-transmission, as these bacteria could become MDR over time.
    \item Conversely, if during the ICU stay a patient has a non-resistant positive culture in other types of bacteria different from the potentially dangerous ones, that patient could coexist with other patients in the ICU, as their germ would not develop resistance over time.
\end{itemize}

\section{Discussion and Conclusions}
\label{sec:Disc}

This study addressed a critical challenge in healthcare: the prediction and interpretability of MDR in ICU patients. We proposed a novel methodology employing a GRU with intrinsic attention mechanisms and post-hoc interpretability techniques, notably the methodological adaptation of SHAP to handle irregular MTS and RNNs with irregular temporal outputs, named IT-SHAP. Our results underscore the clinical relevance and robustness of our approach.

MDR is an increasingly severe clinical issue exacerbated by the improper use of antibiotics~\cite{world2024bacterial}. The proliferation of MDR pathogens threatens the viability of healthcare systems by increasing treatment costs, patient comorbidities, and resource wastage~\cite{whoAntimicrobialResistance_2023, world2024bacterial}. This problem is particularly acute in ICUs due to the vulnerable health status of patients. Longitudinal EHRs, which capture patient health data over time, have proven invaluable for clinical decision support~\cite{challenges_MTS_2022}. However, the heterogeneity and irregularity of MTS data recorded in EHRs present significant analytical challenges~\cite{challenges_MTS_2022}.

Our model's predictive performance, characterized by high initial specificity and increasing sensitivity over time, aligns well with clinical observations. Prolonged ICU stays often result in deteriorating health conditions and higher risks of MDR infections, emphasizing the importance of early identification and management of at-risk patients through interventions such as preventive isolation and tailored treatments. The GRU model with the Hadamard attention mechanism enhances classification performance, maintaining higher specificity and sensitivity during extended ICU stays. Specifically, the GRU model achieved an average sensitivity of 67.07\% ± 1.91\% and a specificity of 68.98\% ± 4.15\%, with an average ROC AUC of 74.08\% ± 2.80\%. In comparison, the GRU model with attention achieved an average sensitivity of 67.91\% ± 3.50\% and a specificity of 77.95\% ± 1.25\%, with an average ROC AUC of 78.27\% ± 1.26\%. While the GRU with attention demonstrated superior specificity and ROC AUC, the sensitivity for classifying MDR patients showed minimal difference between the two models. Compared to previous state-of-the-art results with similar input data, our enhanced model significantly improves both specificity and ROC AUC, highlighting its efficacy in clinical applications.

Our interpretability analysis reveals valuable insights into the factors contributing to MDR. The Hadamard attention mechanism highlights the critical role of prior non-resistant cultures and the influence of antibiotics and co-patient environments in the development of MDR. IT-SHAP further supports these findings by demonstrating the importance of specific bacterial cultures and antibiotic usage patterns. Notably, the presence of non-resistant cultures in dangerous bacteria at ICU admission is a strong predictor of future MDR, emphasizing the need for rigorous infection control measures. Despite the strengths of our approach, the CMI method showed limitations. While CMI provides specific variable importance, it is highly sensitive to noise and parameter selection, potentially affecting the reliability of its conclusions. This sensitivity underscores the need for robust parameter tuning and noise reduction techniques in future research.

The methodology presented in this work could facilitate more intelligent antibiotic treatments and better organization of ICU spaces to reduce germ transmission, potentially preventing MDR germ outbreaks. As EHR data can be extrapolated to any clinical problem, our proposed methodology paves the way for explainable prediction support systems applicable to various clinical issues, expanding the relevance and application of our findings. Furthermore, the proposed methodology can be employed to address other real-world problems modeled as regular or irregular MTS, regardless of the data source.

In conclusion, our proposed methodology offers a significant advancement in predicting and understanding MDR in ICU patients. The combination of GRU with attention mechanisms and IT-SHAP provides a comprehensive tool that balances predictive performance and interpretability, essential for clinical applications. Our findings support the clinical utility of this approach, enabling healthcare professionals to identify at-risk patients early and implement appropriate interventions to mitigate the spread of MDR.

Future research should focus on validating the scalability and generalizability of our methodology using public datasets. Exploring diverse datasets will help to assess the robustness of our approach across different patient populations and healthcare settings. Additionally, integrating advanced noise reduction techniques and further refining parameter tuning processes can enhance the reliability of interpretability methods like CMI. Ultimately, expanding the application of our methodology to various clinical scenarios will contribute to more effective and personalized healthcare interventions.

\section*{Declaration of Competing Interest}

The authors declare that they have no known competing financial interests or personal relationships that could have appeared to influence the work reported in this paper.

\section*{Acknowledgments}
\label{acks}

Work supported by the Spanish federal grants PID2019-105032GB-I00, PID2019-
107768RA-I00 \& PID2022-136887NB-I00 (all funded by the agency AEI/10.13039/501100011033), and the Autonomous Community of Madrid (Madrid ELLIS Unit).


\bibliographystyle{elsarticle-num} 
\bibliography{references}

\begin{thebibliography}{10}
\expandafter\ifx\csname url\endcsname\relax
  \def\url#1{\texttt{#1}}\fi
\expandafter\ifx\csname urlprefix\endcsname\relax\def\urlprefix{URL }\fi
\expandafter\ifx\csname href\endcsname\relax
  \def\href#1#2{#2} \def\path#1{#1}\fi

\bibitem{AI_health_2023}
N.~S. Gupta, P.~Kumar, Perspective of artificial intelligence in healthcare data management: A journey towards precision medicine, Comput. Biol. Med. (2023).

\bibitem{XAI_categories_2022}
G.~Yang, et~al., Unbox the black-box for the medical explainable {AI} via multi-modal and multi-centre data fusion: A mini-review, two showcases and beyond, Inf. Fusion 77 (2022).

\bibitem{XAI_taxonomy_2023}
G.~Schwalbe, B.~Finzel, A comprehensive taxonomy for explainable artificial intelligence: {A} systematic survey of surveys on methods and concepts, Data Min. Knowl. Discov. (2023).

\bibitem{XAI_health_2023}
Y.~C. Wang, et~al., An improved explainable artificial intelligence tool in healthcare for hospital recommendation, Healthc. Anal. 3 (2023).

\bibitem{challenges_MTS_2022}
F.~Xie, et~al., Deep learning for temporal data representation in electronic health records: {A} systematic review of challenges and methodologies, J. Biomed. Inform. 126 (2022).

\bibitem{analysis_RNN_2023}
F.~M. Shiri, T.~Perumal, N.~Mustapha, R.~Mohamed, A comprehensive overview and comparative analysis on deep learning models: {CNN}, {RNN}, {LSTM}, {GRU}, arXiv preprint arXiv:2305.17473 (2023).

\bibitem{murray2022global}
C.~Murray, et~al., Global burden of bacterial antimicrobial resistance in 2019: {A} systematic analysis, The Lancet (2022).

\bibitem{whoAntimicrobialResistance_2023}
{World Health Organization}, Antimicrobial resistance, \url{https://www.who.int/news-room/fact-sheets/detail/antimicrobial-resistance}, [Accessed 27-05-2024] (November 2023).

\bibitem{AMR_deaths_2023}
K.~W.~K. Tang, et~al., Antimicrobial resistance ({AMR}), Br. J. Biomed. Sci. 80 (2023).

\bibitem{covid19_guidelines_2023}
A.~S. Suleiman, et~al., A meta-meta-analysis of co-infection, secondary infections, and antimicrobial resistance in {COVID-19} patients, J. Infect. Public Health 16~(10) (2023).

\bibitem{wartu2019multidrug}
J.~Wartu, et~al., Multidrug resistance by microorganisms: A review, Sci. World J. 14~(4) (2019).

\bibitem{world2024bacterial}
{World Health Organization}, {WHO} bacterial priority pathogens list, 2024: {Bacterial} pathogens of public health importance to guide research, development and strategies to prevent and control antimicrobial resistance, \url{https://www.who.int/publications/i/item/9789240093461}, [Accessed 27-05-2024] (May 2024).

\bibitem{timeshap2021}
J.~Bento, et~al., Timeshap: Explaining recurrent models through sequence perturbations, in: Proc. ACM SIGKDD Conf. Knowl. Discov. Data Min., 2021.

\bibitem{shap2017}
S.~M. Lundberg, S.-I. Lee, A unified approach to interpreting model predictions, in: Proc. Adv. Neural Inf. Process. Syst. 30 (2017).

\bibitem{soydaner2022attention}
D.~Soydaner, Attention mechanism in neural networks: {Where} it comes and where it goes, Neural Comput. Appl. 34~(16) (2022).

\bibitem{figueroa2021towards}
J.~Figueroa~Barraza, et~al., Towards interpretable deep learning: {A} feature selection framework for prognostics and health management using deep neural networks, Sensors 21~(17) (2021).

\bibitem{rw_xai4_2020}
F.~Jimenez, et~al., Feature selection based multivariate time series forecasting: An application to antibiotic resistance outbreaks prediction, Artif. Intell. Med. 104 (2020).

\bibitem{rw_xai3_2023_transAMR}
T.~Mukunthan, et~al., Trans {AMR}: {An} interpretable transformer model for accurate prediction of antimicrobial resistance using antibiotic administration data, IEEE Access (2023).

\bibitem{rw_1_2022}
S.~N. Rich, et~al., Development of a prediction model for antibiotic-resistant urinary tract infections using integrated electronic health records from multiple clinics in {N}orth-{C}entral {F}lorida, Infect. Dis. Ther. 11~(5) (2022).

\bibitem{rw_2_2023}
N.~J. Rhodes, et~al., Machine learning to stratify methicillin-resistant staphylococcus aureus risk among hospitalized patients with community-acquired pneumonia, Antimicrob. Agents Chemother. 67~(1) (2023).

\bibitem{rw_3_2022}
Q.~Liang, et~al., Early prediction of carbapenem-resistant gram-negative bacterial carriage in intensive care units using machine learning, J. Glob. Antimicrob. Resist. 29 (2022).

\bibitem{rw_4_2023}
Y.~Wang, et~al., A deep learning model for predicting multidrug-resistant organism infection in critically ill patients, J. Intensive Care 11~(1) (2023).

\bibitem{rw_5_2022}
R.~Kousovista, et~al., Quantifying the effect of in-hospital antimicrobial use on the development of colistin-resistant acinetobacter baumannii strains: {A} time series analysis, Eur. J. Hosp. Pharm. 29~(2) (2022).

\bibitem{liu2023time}
Z.~Liu, et~al., Time series multi-step forecasting based on memory network for the prognostics and health management in freight train braking system, IEEE Trans. Intell. Transp. Syst. (2023).

\bibitem{guo2023multivariate}
Q.~Guo, et~al., Multivariate time series forecasting using multiscale recurrent networks with scale attention and cross-scale guidance, IEEE Trans. Neural Netw. Learn. Syst. (2023).

\bibitem{2021spatio}
B.~Nikparvar, et~al., Spatio-temporal prediction of the {COVID-19} pandemic in us counties: {Modeling} with a deep {LSTM} neural network, Sci. Rep. 11~(1) (2021).

\bibitem{hernandez2020modelling}
A.~Hern{\'a}ndez~Carnerero, et~al., Modelling temporal relationships in pseudomonas aeruginosa antimicrobial resistance prediction in intensive care unit, in: Proc. AAI4H, Adv. Artif. Intell. Healthc. Med., 2020.

\bibitem{hernandez2021antimicrobial}
{\`A}.~Hern{\`a}ndez-Carnerero, et~al., Antimicrobial resistance prediction in intensive care unit for pseudomonas aeruginosa using temporal data-driven models, Int. J. Interact. Multimedia Artif. Intell. (2021).

\bibitem{martinez2020aplying}
S.~Mart{\'\i}nez-Ag{\"u}ero, et~al., Aplying {LSTM} networks to predict multi-drug resistance using binary multivariate clinical sequences., in: Proc. STAIRS at Europ. Conf. in Artificial Intell., 2020.

\bibitem{escudero2020temporal}
{\'O}.~Escudero-Arnanz, et~al., Temporal feature selection for characterizing antimicrobial multidrug resistance in the intensive care unit., in: Proc. AAI4H at Europ. Conf. in Artificial Intell., 2020.

\bibitem{martinez2022interpretable}
S.~Mart{\'\i}nez-Ag{\"u}ero, et~al., Interpretable clinical time-series modeling with intelligent feature selection for early prediction of antimicrobial multidrug resistance, Future Gener. Comput. Syst. 133 (2022).

\bibitem{martinez2023LSTM}
{\`A}.~Hern{\`a}ndez-Carnerero, et~al., Dimensionality reduction and ensemble of lstms for antimicrobial resistance prediction, Artif. Intell. Med. 138 (2023).

\bibitem{vanishgrad_2021}
S.-H. Noh, Analysis of gradient vanishing of {RNNs} and performance comparison, Information 12~(11) (2021).

\bibitem{GRU_2023}
Z.~Niu, et~al., Recurrent attention unit: A new gated recurrent unit for long-term memory of important parts in sequential data, Neurocomputing 517 (2023).

\bibitem{horn1990hadamard}
R.~A. Horn, The hadamard product, in: Proc. Symp. Appl. Math, Vol.~40, 1990.

\bibitem{gupta2024comparative}
J.~Gupta, K.~Seeja, A comparative study and systematic analysis of {XAI} models and their applications in healthcare, Arch. Comput. Methods Eng. (2024).

\bibitem{ali2023explainable}
S.~Ali, et~al., Explainable artificial intelligence ({XAI}): What we know and what is left to attain trustworthy artificial intelligence, Inf. Fusion 99 (2023).

\bibitem{tradeoff_xai_2020}
A.~B. Arrieta, et~al., Explainable artificial intelligence ({XAI}): Concepts, taxonomies, opportunities and challenges toward responsible {AI}, Inf. Fusion 58 (2020).

\bibitem{gu2022conditional}
X.~Gu, et~al., Conditional mutual information-based feature selection algorithm for maximal relevance minimal redundancy, Appl. Intell. 52~(2) (2022).

\bibitem{zan2022conditional}
L.~Zan, et~al., A conditional mutual information estimator for mixed data and an associated conditional independence test, Entropy 24~(9) (2022).

\bibitem{MI_shannon_2022}
M.~Tisoc, J.~V. Beltr{\'a}n, Mutual information: A way to quantify correlations, Rev. Bras. Ensino F{\'\i}s. 44 (2022).

\bibitem{shannon2_2020}
O.~C. Mesner, C.~R. Shalizi, Conditional mutual information estimation for mixed, discrete and continuous data, IEEE Trans. Inf. Theory 67~(1) (2020).

\bibitem{attention_2022}
T.~Gon{\c{c}}alves, et~al., A survey on attention mechanisms for medical applications: {Are} we moving toward better algorithms?, IEEE Access 10 (2022).

\bibitem{ICU_30Days_2022}
Y.~Deng, et~al., Explainable time-series deep learning models for the prediction of mortality, prolonged length of stay and 30-day readmission in intensive care patients, Front. Med. 9 (2022).

\bibitem{Attention_2019}
D.~A. Kaji, et~al., An attention based deep learning model of clinical events in the intensive care unit, PLoS One 14~(2) (2019).

\bibitem{attention_2017}
J.~Bradbury, et~al., Quasi-recurrent neural networks, in: Proc. Int. Conf. Learn. Represent., 2017.

\bibitem{hinman1992meeting}
A.~R. Hinman, J.~M. Hughes, D.~E. Snider~Jr, M.~L. Cohen, Meeting the challenge of multidrug-resistant tuberculosis: {Summary} of a conference, Morbid. Mortal. Weekly Rep. 41 (1992).

\bibitem{thombley2010menu}
M.~Thombley, D.~Stier, Menu of suggested provisions for state tuberculosis prevention and control laws, US Department of Health and Human Services. Centers for Disease Control and Prevention, Atlanta (2010).

\bibitem{roc_2022}
F.~S. Nahm, Receiver operating characteristic curve: {Overview} and practical use for clinicians, Korean J. Anesthesiol. 75~(1) (2022).

\bibitem{imbalance_2022}
S.~Das, et~al., On supervised class-imbalanced learning: An updated perspective and some key challenges, IEEE Trans. Artif. Intell. 3~(6) (2022).

\bibitem{aurelio2019learning}
Y.~S. Aurelio, et~al., Learning from imbalanced data sets with weighted cross-entropy function, Neural Process. Lett. 50 (2019).

\bibitem{early_stop_2022}
B.~Sabiri, et~al., Mechanism of overfitting avoidance techniques for training deep neural networks., in: Proc. Int. Conf. Enterprise Inf. Syst., 2022.

\bibitem{trubiano2015nosocomial}
J.~A. Trubiano, A.~A. Padiglione, Nosocomial infections in the intensive care unit, Anaesthesia Intensive Care Med. 16~(12) (2015).

\bibitem{kraskov2004estimating}
A.~Kraskov, et~al., Estimating mutual information, Phys. Rev. E 69~(6) (2004).

\end{thebibliography}

\end{document}